\documentclass[journal]{IEEEtran}
\ifCLASSINFOpdf
   \usepackage[pdftex]{graphicx}
\else
   \usepackage[dvips]{graphicx}
\fi
\usepackage{amsmath}
\usepackage{dblfloatfix}

\ifCLASSOPTIONcaptionsoff
 \usepackage[nomarkers]{endfloat}
 \let\MYoriglatexcaption\caption
 \renewcommand{\caption}[2][\relax]{\MYoriglatexcaption[#2]{#2}}
\fi
\usepackage{amssymb}
\usepackage{enumerate}
\usepackage{algorithm}
\usepackage{algcompatible}
\floatname{algorithm}{Pseudocode}

\usepackage{tabularx}
\usepackage{subcaption}
\begin{document}
%
\title{Correlated Parameters to Accurately Measure Uncertainty in Deep Neural Networks}
%
%
%

\author{Konstantin~Posch,
        and~Juergen~Pilz
\thanks{Manuscript received April 19, 2005; revised August 26, 2015.}
\thanks{K. Posch and J. Pilz are with the Department
of Statistics, University Klagenfurt, Austria (e-mails: Konstantin.Posch@aau.at, Juergen.Pilz@aau.at).}
}

%
%

\markboth{Journal of \LaTeX\ Class Files,~Vol.~14, No.~8, August~2015}%
{Shell \MakeLowercase{\textit{et al.}}: Bare Demo of IEEEtran.cls for IEEE Journals}
%



\maketitle

\begin{abstract}
In this article a novel approach for training deep neural networks using Bayesian techniques is presented. The Bayesian methodology allows for an easy evaluation of model uncertainty and additionally is robust to overfitting. These are commonly the two main problems classical, i.e. non-Bayesian, architectures have to struggle with. The proposed approach applies variational inference in order to approximate the intractable posterior distribution. In particular, the variational distribution is defined as product of multiple multivariate normal distributions with tridiagonal covariance matrices. Each single normal distribution belongs either to the weights, or to the biases corresponding to one network layer. The layer-wise a posteriori variances are defined based on the corresponding expectation values and further the correlations are assumed to be identical. Therefore, only a few additional parameters need to be optimized compared to non-Bayesian settings. The novel approach is successfully evaluated on basis of the popular benchmark datasets MNIST and CIFAR-10.
\end{abstract}


%
\IEEEpeerreviewmaketitle

\section{Introduction}
\label{introduction}
%
%
%
%
\IEEEPARstart{N}{owadays}, due to excellent results obtained in many fields of applied machine learning including computer vision and natural language processing \cite{Bengio2006} the popularity of deep learning is increasing rapidly. One of the reasons can surely be found in the fact that Krizhevsky
et al. \cite{Krizhevsky2012} outperformed the competitors in the ImageNet Large Scale Visual Recognition Challenge 2012 by proposing a convolutional neural network (CNN) named AlexNet. While AlexNet includes eight layers, more recent architectures for image classification go even deeper \cite{Simonyan2014VeryDC,He2016DeepRL}. It is well known, that a feed-forward network with merely one hidden layer can approximate a broad class of functions abitrarily well. A mathematical more profound formulation of this statement, the so-called universal approximation theorem, was proven by Hornik et al. \cite{Hornik1990UniversalAO}. However, Liang and Srikant \cite{Liang2016WhyDN} could show that deep nets require exponential less parameters than shallow ones in order to achive a given degree of approximation. Possible applications of deep nets for computer vision include medical imaging, psychology, the automotive industry, finance and life sciences \cite{Jozwik2017,Greenspan2016,Gulshan2016,Banerjee2017,Heaton2016,Li2017}.

Despite the large and ever-increasing number of real world use cases deep learning comes along with two restrictions which still limit its areas of application. The first restriction is that deep networks require a large amount of training data, otherwise they are prone to overfiting. The reason for this is the huge amount of parameters neural nets hold. Although deep nets require exponential less parameters than shallow ones, the remaining number is nevertheless very high. Thus, in many potential fields of application, where such an amount cannot be provided, deep learning is of limited use, or often even cannot be used. To counteract this problem commonly diverse regularization techniques are applied. Besides classical approaches, such as the penalization of the L2 norm or the L1 norm, stochastic regularization methods gain increasing attention. For instance, dropout \cite{dropout} and dropconnect \cite{dropconnect} count to these stochastic techniques. The first one randomly sets the activation of non-output neurons to zero during network training and the second one randomly sets network weights to zero. While dropout classically is interpreted as an efficient way of performing model averaging with neural networks, Gal
and Ghahramani \cite{Gal2015orig} as well as Kingma et al. \cite{Kingma2015} recently showed that it can also be considered as an application of Bayesian statistics. The second restriction deep networks struggle with is that prediction uncertainty cannot be measured. Especially, in the medical field or for self-driving vehicles it is essential that the prediction uncertainty can be determined \cite{Roman2017}. In these areas of application a model which predicts on average quite well is not good enough. One has to know if the model is certain in its predictions or not, such that in the case of high uncertainty a human can decide instead of the machine. Please be aware of the fact that the probabilites obtained when running a deep net for a classification task should not be interpreted as the confidence of the model. As a matter of fact a neural net can guess randomly while returning a high class probability \cite{GalUnc}.

A possible strategy to overcome the restrictions classical deep learning has to deal with is applying Bayesian statistics. In so-called Bayesian deep learning the network parameters are treated like random variables and are not considered to be fixed deterministic values. In particular, an a priori distribution is assigned to them and updating the prior knowledge after observing traning data results in the so-called posterior distribution. The uncertainty in the network parameters can be directly translated in uncertainty about predictions. Further, Bayesian methods are robust to overfitting because of the built-in regularization due to the prior. Buntine and S. Weigend \cite{Buntine1991} were one of the first who presented approximate Bayesian methods for neural nets. Two years later Hinton and van Camp \cite{Hinton1993KeepingTN} already proposed the first variational methods. Variational methods try to approximate the true posterior distribution with another parameteric distribution, the so-called variational distribution. The approximation takes place due to an optimization of the parameters of the variational distribution. They followed the idea that there should be much less information in the weights than in the output vectors of the training cases in order to allow for a good generalization of neural networks. Denker and Lecun \cite{Denker91transformingneural-net} as well as J.C. MacKay \cite{MacKay1995} used Laplace approximation in order to investigate the posterior distributions of neural nets. Neal \cite{neal2012bayesian} proposed and investigated hybrid Monte Carlo training for neural networks as a less limited alternative to the Laplace approximation. However, the approaches mentioned up to now are often not scalable for modern applications which go along with highly parameterized networks. Graves \cite{Graves2011PracticalVI} was the first to show how variational inference can be applied to modern deep neural networks due to Monte Carlo integration. He used a Gaussian distribution with a diagonal covariance matrix as variational distribution. Blundell et al. \cite{Blundell:2015:WUN:3045118.3045290} extended and improved the work of Graves \cite{Graves2011PracticalVI} and also used a diagonal Gaussian to approximate the posterior. As already mentioned, Gal and Ghahramani \cite{Gal2015orig} as well as Kingma et al. \cite{Kingma2015} showed that using the regularization technique dropout can also be considered as variational inference.

The indepenece assumptions going along with variational inference via diagonal Gaussians (complete independence of network parameters), or also going along with variational inference according to dropout (independence of neurons) are restrictive. Permitting an exchange of information between different parts of neural network architectures should lead to more accurate uncertainty estimates. Louizos and Welling \cite{Louizos2016} used a distribution over random matrices in order to define the variational distribution. Thus, they could reduce the amount of variance-related parameters that have to be estimated and further allow for an information sharing between the network weights. Note that in the diagonal Gaussian approach assigning one variance term to each random weight and one variance term to each random bias doubles the amount of parameters to optimize in comparison to frequentist deep learning. Consequently, network training becomes more complicated and computationally expensive.

It should be mentioned that variational Bayes is just
a specific case of local $\alpha$-divergence minimization. According to Amari \cite{Amari1985} the $\alpha$-divergence between two densities $p(\mathbf{w})$ and $q(\mathbf{w})$ is given by $D_{\alpha}(p(\mathbf{w})||q(\mathbf{w}))=\frac{1}{\alpha(1-\alpha)}\left(1-\int p(\mathbf{w})^{\alpha}q(\mathbf{w})^{(1-\alpha)}~d\mathbf{w}\right)$. Thus, the $\alpha$-divergence converges for $\alpha\rightarrow 0$ to the KL divergence \cite{kullback1951} which is typically used in variational Bayes. It has been shown \cite{Hernndez2016,GalAlpha2017} that an optimal choice of $\alpha$ is task specific and that non-standard settings, i.e. settings with $\alpha\neq 0$ can lead to better prediction results and uncertainty estimates. 

However, in this paper we do not propose an optimal choice of $\alpha$. Rather, for the classical case $\alpha=0$ we will propose a good and easy to interpret variational distribution. For this task recent work from Posch et al. \cite{Posch2019} is extended. They used a product of  Gaussian distributions with specific diagonal covariance matrices in order to define the variational distribution. In particular, the a posteriori uncertainty of the network parameters is represented per network layer and depending on the estimated parameter expectation values. Therefore, only few additional parameters have to be optimized compared to classical deep learning and the parameter uncertainty itself can easily be analyzed per network layer. We extened this distribution by allowing network parameters to be correlated with each other. In particular, the diagonal covariance matrices are replaced with tridiagonal ones. Each tridiagonal matrix is defined in such a way that the correlations between neighboured parameters are identical. This way of treating network layers as units in terms of dependence allows for an easy analysis of the dependence between network parameters. Moreover, again only few additional parameters compared to classical deep learning need to be optimized, which guarantees that the difficulty of the network optimization does not increase significantly. Note that our extension allows for an exchange of information between different parts of the network and therefore should lead to more reliable uncertainty estimates. We have evaluated our approach on basis of the popular benchmark datasets MNIST \cite{Deng2012} and CIFAR-10 \cite{Krizhevsky2009LearningML}.  The promising results can be found in Section \ref{experiments}.

\section{Background}
\label{background}

In this section based on previous work \cite{Hinton1993KeepingTN,Graves2011PracticalVI,GalUnc,Bishop2006,Posch2019}  we briefly discuss how variational inference can be applied in deep learning. Since we are particularly interested in image classification the focus will be on networks designed for classification tasks. Note that the methodology also holds for regression models after some slight modifications.

Let $\mathbf{W}$ denote the random vector covering all parameters (weights and biases) of a given neural net $\mathbf{f}$. Further, let $p(\mathbf{w})$ denote the density used to define a priori knowledge regarding $\mathbf{W}$. According to the Bayesian theorem the posterior distribution of $\mathbf{W}$  is given by the density
$$p(\mathbf{w}|\mathbf{y},\mathbf{X})=\frac{p(\mathbf{y}|\mathbf{w},\mathbf{X})p(\mathbf{w})}{\int p(\mathbf{y}|\mathbf{w},\mathbf{X})p(\mathbf{w})~d\mathbf{w}}$$
where $\mathbf{X} = \{\mathbf{x}_1,...,\mathbf{x}_{\beta}\}$ denotes a set of
training examples and $\mathbf{y}=(y_1,...,y_{\beta})^T$ holds the corresponding class labels. Note that the probability $p(\mathbf{y}|\mathbf{w},\mathbf{X})$ is given by the product $\prod_{i=1}^{\beta}\mathbf{f}(\mathbf{x}_i;\mathbf{w})_{y_i}$ in accordance with the classical assumptions on stochastic independence and modeling in deep learning for classification. The integral in the above representation of $p(\mathbf{w}|\mathbf{y},\mathbf{X})$ is commonly intractable due to its high dimension $\beta$. Variational inference aims at approximating the posterior with another parametric distribution, the so-called variational distribution $Q_{\boldsymbol\phi}(\mathbf{w})$. To this end the so-called variational parameters $\boldsymbol\phi$ are optimized by minimizing the Kullback Leibler divergence (KL divergence)
$$D_{KL}(q_{\boldsymbol\phi}(\mathbf{w})||p(\mathbf{w}|\mathbf{y},\mathbf{X}))=\mathbb{E}_{q_{\boldsymbol\phi}(\mathbf{w})}\left(\operatorname{ln}\frac{q_{\boldsymbol\phi}(\mathbf{w})}{p(\mathbf{w}|\mathbf{y},\mathbf{X})}\right)$$
between the variational distribution and the posterior. Although the KL divergence is no formal distance measure (does not satisfy some of the requested axioms) it is a common choice to measure the similarity of probability distributions.

Since the posterior distribution is unknown the divergence $D_{KL}(q_{\boldsymbol\phi}(\mathbf{w})||p(\mathbf{w}|\mathbf{y},\mathbf{X}))$ cannot be minimized directly. Anyway, the minimization of $D_{KL}(q_{\boldsymbol\phi}(\mathbf{w})||p(\mathbf{w}|\mathbf{y},\mathbf{X}))$ is equivalent to the minimization of the so-called negative log evidence lower bound
\begin{align*}
L_{VI}&=-\mathbb{E}_{q_{\boldsymbol\phi}(\mathbf{w})}\left[\operatorname{ln}p(\mathbf{y}|\mathbf{w},\mathbf{X})\right]+D_{KL}(q_{\boldsymbol\phi}(\mathbf{w})||p(\mathbf{w}))\\
&=-\sum\limits_{i=1}^{\beta}\left\{\mathbb{E}_{q_{\boldsymbol\phi}(\mathbf{w})}\left[\operatorname{ln}\mathbf{f}(\mathbf{x}_i;\mathbf{w})_{y_i}\right]\right\}+D_{KL}(q_{\boldsymbol\phi}(\mathbf{w})||p(\mathbf{w})).
\end{align*}
Thus, the optimization problem reduces to the minimization of the sum of the expected negative log likelihood and the KL divergence between the variational distribution and the prior with respect to $\boldsymbol\phi$. Inspired by stochastic gradient descent it is not unusual to approximate the expected values in $L_{VI}$ via Monte Carlo integration with one sample during network training. Note that the re-sampling in each training iteration guarantees that a sufficient amount of samples is drawn overall. Commonly, mini-batch gradient descent is used for optimization in deep learning. To take account of the resulting reduction of the number of training examples used in each iteration of the optimization the objective function has to be rescaled. Thus, in the  $k$-th iteration the function to minimize is given by
$$L_{VI}^k=-\frac{1}{m}\sum\limits_{i=1}^{m}\left\{\operatorname{ln}\mathbf{f}(\tilde{\mathbf{x}}_i;\mathbf{w}_k)_{\tilde{y}_i}\right]\}+\frac{1}{\beta}D_{KL}(q_{\boldsymbol\phi}(\mathbf{w})||p(\mathbf{w}))$$
where $\mathbf{w}_k$ denotes a sample from $q_{\boldsymbol\phi}(\mathbf{w})$, $m$ denotes the mini-batch size, and $\tilde{\mathbf{x}}_1,...,\tilde{\mathbf{x}}_m,\tilde{y}_1,...,\tilde{y}_m$ denote the mini-batch itself.

Summing up, optimizing a Bayesian neural net is quite similar to the optimization of a classical one. While in frequentist deep learning it is common to penalize the Euclidean norm of the network parameters in terms of regularization in Bayesian deep learning deviations of the variational distribution from the prior are penalized. In principal, the same loss function $L$ (cross entropy loss) is minimized, but with the crucial difference that network parameters have to be sampled since they are random.

In Bayesian deep learning predictions are based on the posterior predictive distribution, i.e. the distribution of a class label $y^*$ for a given example $\mathbf{x}^*$ conditioned on the observed data $\mathbf{y},\mathbf{X}$. The distribution can be approximated via Monte Carlo integration
\begin{align*}
p(y^*|\mathbf{x}^*,\mathbf{y},\mathbf{X})&=\int p(y^*|\mathbf{w},\mathbf{x}^*)p(\mathbf{w}|\mathbf{y},\mathbf{X})~d\mathbf{w}\\
&\approx\int p(y^*|\mathbf{w},\mathbf{x}^*)q_{\boldsymbol\phi}(\mathbf{w})~d\mathbf{w}\\
&\approx\frac{1}{N}\sum\limits_{i=1}^{N}\mathbf{f}(\mathbf{x}^*;\mathbf{w}_i)_{y^*}
\end{align*}
where $\mathbf{w}_1,...,\mathbf{w}_N$ denote samples from $q_{\boldsymbol\phi}(\mathbf{w})$. Therefore, the class of an object $\mathbf{x}^*$ is predicted by computing multiple network outputs with parameters sampled from the variational distribution. Averaging the output vectors results in an estimate of the posterior predictive distribution, such that the a posteriori most probable class finally serves as prediction.

The a posteriori uncertainty in the network parameters $\mathbf{W}$ can directly be translated in uncertainty about the random network output $\mathbf{f}(\mathbf{x}^*;\mathbf{W})$ and thus the posterior probability of an object $\mathbf{x}^*$ belonging to class $y^*$. Therefore, at first multiple samples $\mathbf{w}_1,...,\mathbf{w}_N$ are drawn from the variational distribution $Q_{\boldsymbol\phi}(\mathbf{w})$. Then the corresponding network outputs $\mathbf{f}(\mathbf{x}^*;\mathbf{w}_1)_{y^*},...,\mathbf{f}(\mathbf{x}^*;\mathbf{w}_n)_{y^*}$ are determined. Finally, the empirical $\frac{\alpha}{2}$ and $(1-\frac{\alpha}{2})$ quantiles of these outputs provide an estimate of the $(1-\alpha)$ credible interval for the probability $p(y^*|\mathbf{x}^*,\mathbf{y},\mathbf{X})$.

\section{Methodology}
\label{methodology}
In this section we describe our novel approach. At first we give a formal definition of the variational distribution we use to approximate the posterior and, additionally, we propose a normal prior. Moreover, we report the derivatives of the approximation $L_{VI}^k$ of the negative log evidence lower bound (see Section \ref{background}) with respect to the variational parameters, i.e. the learnable parameters. Finally, we present the pseudocode which is the basis of our implementation of the proposed method. 

\subsection{Variational Distribution and Prior}
\label{variationaldist}
Let $\mathbf{W}_j=(W_{j1},...,W_{jK_j})^T$ denote the random weights of layer $j$ $(j=1,...,d)$. Further, let $\mathbf{B}_j=(B_{j1},...,B_{jk_j})^T$ denote the corresponding random bias terms. The integers $K_j$ and $k_j$ denote the number of weights and the number of biases in layer $j$, respectively.

As already mentioned in Section \ref{introduction}, we define the variational distribution as a product of multivariate normal distributions with tridiagonal covariance matrices. Applying variational inference to Bayesian deep learning presupposes that samples from the variational distribution can be drawn during network training as well as at the stage in which new samples are used for prediction. Especially at the training phase, it is essential that the random sampling can be reduced to the sampling from a (multivariate) standard normal distribution and appropriate affine-linear transformation of the drawn samples based on the learnable parameters. A direct sampling from the non-trivial normal distributions would mask the variational parameters and thus make it impossible to optimize them by gradient descent. Provided that a covariance matrix is positive definite (in general covariance matrices are only positive semidefinte) there exists a unique Cholesky decomposition of the matrix which can be used for this task. Note that for each real-valued symmetric positive-definite square matrix a unique decomposition (Cholesky decomposition) of the form $\mathbf{\Sigma}=\mathbf{L}\mathbf{L}^T$ exists, where L is a lower triangular matrix with real and positive diagonal entries. Thus, we are interested in symmetric tridiagonal matrices which always stay positive definite no matter how the corresponding learnable parameters are adjusted during network training. The first thing required is a criterion for the positive definitness of for our purposes approriate tridiagonal matrices. Andeli\'c and Fonseca \cite{Andelic2011} gave the following sufficient condition for positive definiteness of tridiagonal matrices:  Let 
$$\mathbf{\Sigma}=\begin{pmatrix}
a_1  & c_1  &        &        &    \\
c_1  & a_2  & c_2    &        &    \\
     & c_2  & a_3    & \ddots &    \\
     &      & \ddots & \ddots &    c_{n-1}\\
     &      &        & c_{n-1}    & a_n\\
\end{pmatrix}\in\mathbb{R}^{n\times n}
$$
a symmetric tridiagonal matrix with positive diagonal entries. If 
\begin{align}
c_i^2<\frac{1}{4}a_ia_{i+1}\frac{1}{\cos^2\left(\frac{\pi}{n+1}\right)}~~~i=1,...,n-1
\end{align} 
then $\mathbf{\Sigma}$ is positive definite. Consider now a matrix $\mathbf{\Sigma}\in\mathbb{R}^{n\times n}$ defined as follows:
\begin{small}
\begin{align*}
\mathbf{\Sigma}&:=\mathbf{L}\mathbf{L}^T\\
&=\begin{pmatrix}
a_1  &      &        &        &    \\
c_1  & a_2  &        &        &    \\
     & c_2  & a_3    &        &    \\
     &      & \ddots & \ddots &    \\
     &      &        & c_{n-1}    & a_n\\
\end{pmatrix}\begin{pmatrix}
a_1  & c_1     &        &        &    \\
  & a_2  &  c_2      &        &    \\
     &   & a_3    &  \ddots      &    \\
     &      &  & \ddots & c_{n-1}   \\
     &      &        &     & a_n\\
\end{pmatrix}\\
&=\begin{pmatrix}
a_1^2  & c_1a_1     &        &        &    \\
c_1a_1  & c_1^2+a_2^2  & c_2a_2       &        &    \\
     & c_2a_2  & c_2^2+a_3^2    & c_3a_3       &    \\
     &     &\vdots &\      &    \\
     &		&	c_{n-2}a_{n-2}	 &	c_{n-2}^2+a_{n-1}^2	  &  c_{n-1}a_{n-1}\\
     &      &        & c_{n-1}a_{n-1}    & c_{n-1}^2+a_n^2\\
\end{pmatrix}
\end{align*}
\end{small}
If $\mathbf{\Sigma}$ satisfies condition $(1)$ and has positive diagonal entries,  the matrix $\mathbf{L}$ defines its Cholesky decomposition and further $\mathbf{\Sigma}$ is a valid covariance matrix since every real, symmetric, and positive semidefinite square matrix defines a valid covariance matrix. As in the work of Posch et al. \cite{Posch2019} we define the variances as multiples of the corresponding expectation values, denoted by $m_1,...,m_n\in\mathbb{R}\setminus\{0\}$
\begin{align}
c_{i-1}^2+a_i^2&:=\tau^2m_i^2~~~i=1,...,n\\
c_0&:=0
\end{align}
where $\tau\in\mathbb{R}^{+}$. Defining the variances proportional to the expectation values allows for a useful specification of them. This specification requires, besides the expectation values, only one additional variational parameter. Moreover, we want the correlations to be identical, which leads to the following covariances $c_ia_i$
\begin{align}
&\rho= \frac{c_ia_i}{\sqrt{\tau^2m_i^2}\sqrt{\tau^2m_{i+1}^2}}=\frac{c_ia_i}{\tau^2|m_i||m_{i+1}|}\\
\Leftrightarrow ~& c_ia_i=\rho \tau^2|m_i||m_{i+1}|
\end{align}
for $i=1,...,n-1$. By rearranging Equations $(2)$ and $(5)$ one obtains a recursive formula for the elements of the matrix $L$:
\begin{align}
(5)\Leftrightarrow a_i &= \frac{\rho \tau^2|m_i||m_{i+1}|}{c_i}
\end{align}
\begin{align}
(2)&\Leftrightarrow c_{i-1}^2 + \frac{\rho^2 \tau^4 m_i^2 m_{i+1}^2}{c_i^2}=\tau^2m_i^2\\
&\Leftrightarrow c_i^2 = \frac{\rho^2 \tau^4 m_i^2 m_{i+1}^2}{\tau^2m_i^2-c_{i-1}^2}
\end{align}
Note that Equation $(8)$ for instance is satisfied for
\begin{align}
c_i= \frac{\rho \tau^2 m_i m_{i+1}}{\sqrt{\tau^2m_i^2-c_{i-1}^2}}.
\end{align} By defining the $c_i$ this way one does not end up by the Cholseky decomposition which assumes the diagonal elements $a_i$ of $L$ to be positive and thus by $(6)$ also assumes the $c_i$ to be positive. Taking the absolute values of the $c_i$ according to Equation $(9)$ would result in the Cholesky decomposition, but is not necessary for our purposes and therefore not done. Thus, the elements of $\mathbf{L}$ are recursively defined as
\begin{align}
c_0&:=0\\
c_i&:=\frac{\rho \tau^2 m_i m_{i+1}}{\sqrt{\tau^2m_i^2-c_{i-1}^2}}~~~i=1,...,n-1\\
a_i &:= \frac{\rho \tau^2|m_i||m_{i+1}|}{c_i}~~~i=1,...,n-1\\
a_n&:=\sqrt{\tau^2m_n^2-c_{n-1}^2}.
\end{align}
Note that the matrix $\mathbf{\Sigma}$ defined by the Equations $(10-13)$  satisfies condition $(1)$ iff
\begin{align}
&(\rho\tau^2|m_i||m_{i+1}|)^2<\frac{1}{4}\tau^2m_i^2\tau^2m_{i+1}^2\frac{1}{\cos^2\left(\frac{\pi}{n+1}\right)}\\
\Leftrightarrow & ~\rho^2 < \frac{1}{4}\frac{1}{\cos^2\left(\frac{\pi}{n+1}\right)}\\
\Leftrightarrow & ~\rho \in\underbrace{\left(-\frac{1}{2}\frac{1}{\sqrt{\cos^2\left(\frac{\pi}{n+1}\right)}},\frac{1}{2}\frac{1}{\sqrt{\cos^2\left(\frac{\pi}{n+1}\right)}}\right)}_{\underset{\text{for large }n}{\approx}\left(-\frac{1}{2},\frac{1}{2}\right)}.
\end{align}
Thus, provided that condition $(16)$ holds there exists a unique Cholesky decomposition of $\mathbf{\Sigma}$ which again guarantees that the $c_i$ according to $(11)$ are well defined and, further, also that $a_n$ according to $(13)$ is well defined.

Using the considerations above the variational distribution can finally be defined as follows. In a first step lower bidiagonal matrices $\mathbf{L}_j$ are specified for $j=1,...,d$:
\begin{small}
\begin{align}
\mathbf{L}_j:=\begin{pmatrix}
a_{j1}  &      &        &        &    \\
c_{j1}  & a_{j2}  &        &        &    \\
     & c_{j2}  & a_{j3}    &        &    \\
     &      & \ddots & \ddots &    \\
     &      &        & c_{j,K_j-1}    & a_{jK_j}\\
\end{pmatrix}
\end{align}
\end{small}
\begin{align}
c_{j0}&:=0\\
c_{ji}&:=\frac{\rho_j \tau_j^2 m_{ji} m_{j,i+1}}{\sqrt{\tau_j^2m_{ji}^2-c_{j,i-1}^2}}~~~i=1,...,K_{j}-1\\
a_{ji} &:= \frac{\rho_j \tau_j^2|m_{ji}||m_{j,i+1}|}{c_{ji}}~~~i=1,...,K_{j}-1\\
a_{jK_j}&:=\sqrt{\tau_j^2m_{jK_j}^2-c_{j,K_j-1}^2}
\end{align}
The variational distribution of the weights of the $j$-th layer is then defined as a multivariate normal distribution
$$Q_{\boldsymbol\phi_j}(\mathbf{w}_j)=\mathcal{N}(\mathbf{w}_j;\mathbf{m}_j,\mathbf{\Sigma}_j)$$
with expected value $\mathbf{m}_j$ and a tridiagonal covariance matrix $\mathbf{\Sigma}_j=\mathbf{L}_j\mathbf{L}_j^T$. According to the considerations above, the variances of the normal distribution are given by $\tau_j^2m_{ji}^2$ ($i=1,...,K_j$) and the covariances are given by $\rho_{j}\tau_{j}^2|m_{ji}||m_{j,i+1}|$ ($i=1,...,K_j-1$). This again implies that the correlations are all the same and given by the parameter $\rho_j$. Since the parameter $\tau_j$ regulates the variances of the distribution it should not take negative values during optimization. To guarantee this it is reparameterized with help of the softplus function
\begin{align}
\tau_j&=\operatorname{ln}(1+\operatorname{exp}(\delta_j))>0.
\end{align}
Moreover, the parameter $\rho_j$ should lie in the interval $(-\frac{1}{2},\frac{1}{2})$ to ensure that the matrix $\mathbf{\Sigma}_j$ is positive definite. In deep learning dimensions are commonly high, such that the approximation in Equation $(16)$ can be considered as valid. The following reparameterization ensures that the desired property holds:
\begin{align}
\rho_j&=\frac{1}{1+\operatorname{exp}(-\gamma_j)}-\frac{1}{2}\in\left(-\frac{1}{2},\frac{1}{2}\right)
\end{align}
In addition, the diagonal entries of $\mathbf{\Sigma}_j$ have to be non-zero to ensure positive definiteness, which again implies that each component of $\mathbf{m}_j$ has to be non-zero. We decide to set values which are not significantly different from zero to small random numbers in our implementation instead of introducing another reparameterization. Finally, $\mathbf{m}_j\in\mathbb{R}^{K_j}\setminus\{\mathbf{0}\}$, $\delta_jþ\inþ\mathbb{R}$, and $\gamma_j\in\mathbb{R}$ can be summarized as the variational parameters $\boldsymbol\phi_j$ corresponding to the weights of the $j$-th network layer.

One can easily sample from a random vector $\mathbf{W}_j$ belonging to this distribution using samples from a standard normal distribution $\mathcal{N}(0,1)$ since it can be written as
\begin{align}
\mathbf{W}_j&= \mathbf{m}_j+\mathbf{L}_j\mathbf{X}_j~~~\text{with}~~~\mathbf{X}_j\sim\mathcal{N}(\mathbf{0}_{K_j},\mathbf{I}_{K_j}).
\end{align}
Note that Equation $(24)$ can also be written as:
\begin{align}
W_{j1}&= m_{j1}+a_{j1}X_{j1}\\
\vdots\nonumber\\
W_{ji}&=m_{ji}+c_{j,i-1}X_{j,i-1}+a_{ji}X_{ji}\\
\vdots\nonumber\\
W_{jK_j}&=m_{jK_j}+c_{j,K_j-1}X_{j,K_j-1}+a_{jK_j}X_{jK_j}
\end{align}

The layer-wise variational distributions of the bias terms denoted by $q_{\boldsymbol\phi_{bj}}(\mathbf{b}_j)$ are defined completely analogous to those of the weights. Assuming independence of the layers as well as independence between weights and biases, the overall variational distribution is given by
$$q_{\boldsymbol{\phi}}(\mathbf{w})=\prod\limits_{j=1}^dq_{\boldsymbol{\phi}_j}(\mathbf{w}_j)q_{\boldsymbol{\phi}_{bj}}(\mathbf{b}_{j})$$
where $\boldsymbol{\phi}_j=\{\mathbf{m}_j$, $\delta_j,\gamma_j\},\boldsymbol{\phi}_{bj}=\{\mathbf{m}_{bj},\delta_{bj},\gamma_{bj}\}$, $q_{\boldsymbol{\phi}_j}(\mathbf{w}_j)$ denotes the density of $\mathcal{N}(\mathbf{m}_j,\mathbf{\Sigma}_j)$, $q_{\boldsymbol{\phi}_{bj}}(\mathbf{b}_{j})$ denotes the density of $\mathcal{N}(\mathbf{m}_{bj},\mathbf{\Sigma}_{bj})$, and $\mathbf{w}$ is a vector including all weights and all biases.

We define the a priori distribution completely analogous to Posch et al. \cite{Posch2019}. In particular, its density is given by:
$$p(\boldsymbol{w})=\prod\limits_{j=1}^dp(\boldsymbol{w}_j)p(\boldsymbol{b}_j)$$
where $p(\boldsymbol{w}_j)$ denotes the density of $N(\boldsymbol{\mu}_j,\zeta_j^2\mathbf{I}_{K_j})$ and $p(\boldsymbol{b}_j)$ denotes the density of $N(\boldsymbol{\mu}_{bj},\zeta_{bj}^2\mathbf{I}_{k_j})$.

\subsection{Kullback Leibler Divergence}
\label{kullback}

The fact that the variational distribution as well as the prior factorize simplifies the computation of the Kullback Leibler divergence. Indeed, the overall divergence is given by the sum of the layer-wise divergences (for further details refer to Posch et al. \cite{Posch2019}):
\begin{align*}
D_{KL}(q_{\boldsymbol{\phi}}(\mathbf{w})||p(\mathbf{w}))&=\sum\limits_{j=1}^d\left[D_{KL}(q_{\boldsymbol{\phi}_{j}}(\mathbf{w}_{j})||p(\mathbf{w}_j))\right]\\
&+\sum\limits_{j=1}^d\left[D_{KL}(q_{\boldsymbol{\phi}_{bj}}(\mathbf{b}_j)||p(\mathbf{b}_j))\right]
\end{align*}
Thus, computing the overall divergence can be reduced to compute $D_{KL}(q_{\boldsymbol{\phi}_{j}}(\mathbf{w}_{j})||p(\mathbf{w}_j))$ for fixed $j\in\{1,...,K_j\}$, since the remaining divergences compute completely analogously (only the indices differ).
According to Hershey and Olsen \cite{Hershey2007} the KL divergence between two $p$-dimensional normal distributions, given by $H(\mathbf{x})=\mathcal{N}(\mathbf{x};\boldsymbol{\mu}_h,\mathbf{\Sigma}_h)$ and $G(\mathbf{x})=\mathcal{N}(\mathbf{x};\boldsymbol{\mu}_g,\mathbf{\Sigma}_g)$, computes as
\begin{align}
D_{KL}(H||G)=\frac{1}{2}&\left[\operatorname{ln}\frac{|\mathbf{\Sigma}_g|}{|\mathbf{\Sigma}_h|}+\operatorname{tr}(\mathbf{\Sigma}_g^{-1}\mathbf{\Sigma}_h)-p\right.\\
&+\left.(\boldsymbol{\mu}_h-\boldsymbol{\mu}_g)^T\mathbf{\Sigma}_g^{-1}(\boldsymbol{\mu}_h-\boldsymbol{\mu}_g)\right].\nonumber
\end{align}
Thus, the determinant of the covariance matrix $\mathbf{\Sigma}_j$ is required for the computation of the Kullback Leibler divergence $D_{KL}(q_{\boldsymbol{\phi}_{j}}(\mathbf{w}_{j})||p(\mathbf{w}_j))$. Using basic properties of determinants $|\mathbf{\Sigma}_j|$ computes as follows for fixed $j$:
\begin{align}
|\mathbf{\Sigma}_j|=|\mathbf{L}_j\mathbf{L}_j^T|=|\mathbf{L}_j||\mathbf{L}_j^T|=|\mathbf{L}|^2=\prod\limits_{i=1}^{K_j}a_{ji}^2
\end{align}
Using $(28)$ and $(29)$ $D_{KL}(q_{\boldsymbol{\phi}_{j}}(\mathbf{w}_{j})||p(\mathbf{w}_j))$ then reads
\begin{small}
\begin{align}
&D_{KL}(q_{\boldsymbol{\phi}_{j}}(\mathbf{w}_{j})||p(\mathbf{w}_j))=\frac{1}{2}\left[\operatorname{ln}\frac{|\zeta_j^2\mathbf{I}_{K_j}|}{|\mathbf{\Sigma}_j|}+ \operatorname{tr}\left((\zeta_j^2\mathbf{Iþ}_{K_j})^{-1}\mathbf{\Sigma}_j\right)\right.\nonumber\\
&\hspace{2.2cm}-\left.K_j+(\mathbf{m}_j-\boldsymbol{\mu}_j)^T\left(\zeta_j^2\mathbf{I}_{K_j}\right)^{-1}(\mathbf{m}_j-\boldsymbol{\mu}_j)\right]\nonumber\\
&=\frac{1}{2}\left[-\operatorname{ln}\left(\prod\limits_{i=1}^{K_j}a_{ji}^2\right)+\frac{\tau_j^2}{\zeta_j^2}||\mathbf{m}_j||_2^2+\frac{1}{\zeta_j^2}||\mathbf{m}_j-\boldsymbol{\mu}_j||_2^2+c\right]\nonumber\\
&=\frac{1}{2}\left[-\sum\limits_{i=1}^{K_j}\left(\operatorname{ln}a_{ji}^2\right)+\frac{\tau_j^2}{\zeta_j^2}||\mathbf{m}_j||_2^2+\frac{1}{\zeta_j^2}||\mathbf{m}_j-\boldsymbol{\mu}_j||_2^2+c\right]
\end{align}
\end{small}
where $c$ always denotes an additive constant.

\subsection{Derivatives}
\label{derivatives}
Commonly neural networks are optimized via mini-batch gradient descent. Thus, in order to train a neural net $\mathbf{f}(\cdot,\mathbf{w})$ according to our novel approach the partial derivatives of the approximation $L_{VI}^k$ of the negative log evidence lower bound described in Section \ref{background} with respect to the variational parameters $\boldsymbol{\phi}_j=\{\mathbf{m}_j$, $\delta_j,\gamma_j\},\boldsymbol{\phi}_{bj}=\{\mathbf{m}_{bj},\delta_{bj},\gamma_{bj}\}$ are required. In particular, the partial derivatives of the loss function $L$ typically used in deep learning and the partial derivatives of the Kullback Leibler divergence between prior and variational distribution have to be computed. Note, that the loss function equals the negative log likelihood of the data and is given by the cross-entropy loss in the case of classification and by the Euclidean loss in the case of regression. Thus, $L$ depends on the network $\mathbf{f}$ itself, with parameters sampled from the variational distribution $q_{\boldsymbol\phi}(\mathbf{w})$. With the help of the multivariate chain rule the required partial derivatives of $L$ can be computed based on the classical derivatives used in non-Bayesian deep learning:
\begin{small}
\begin{align}
\frac{\partial L}{\partial  \mathbf{m}_j}=\left(\frac{\partial \mathbf{w}_j}{\partial  \mathbf{m}_j}\right)^T\frac{\partial L}{\partial  \mathbf{w}_j}~~~&\Rightarrow~~~\frac{\partial L}{\partial m_{ji}}=\sum\limits_{l}\frac{\partial L}{\partial w_{jl}}\frac{\partial w_{jl}}{\partial m_{ji}}\\
\frac{\partial L}{\partial \delta_j}=\left(\frac{\partial \mathbf{w}_j}{\partial \delta_j}\right)^T\frac{\partial L}{\mathbf{w}_j}~~~&\Rightarrow~~~\frac{\partial L}{\delta_j}=\sum\limits_l\frac{\partial L}{\partial w_{jl}}\frac{\partial w_{jl}}{\partial \delta_j}\\
\frac{\partial L}{\partial \gamma_j}=\left(\frac{\partial \mathbf{w}_j}{\partial \gamma_j}\right)^T\frac{\partial L}{\mathbf{w}_j}~~~&\Rightarrow~~~\frac{\partial L}{\gamma_j}=\sum\limits_l\frac{\partial L}{\partial w_{jl}}\frac{\partial w_{jl}}{\partial \gamma_j}
\end{align}
\end{small}
Equations $(31-33)$ only deal with the derivatives of $L$ with respect to the variational parameters belonging to the network weights. Completely analogous equations hold for the bias terms. In the sequel we focus on the derivatives of the weights, since the derivatives for the biases are obviously of the same form. Note that for a given sample $\mathbf{w}$ from the variational distribution the layer-wise derivatives $\frac{\partial L}{\partial  \mathbf{w}_j}, \frac{\partial L}{\partial  \mathbf{b}_j}$ ($j=1,...,d$) are computed as in non-Bayesian deep learning. Thus, the problem of finding closed form expressions for the required derivatives of $L$ reduces to the problem of finding these expressions for the $\mathbf{w}_j$'s and the $\mathbf{b}_j$'s. Taking account of the Equations $(25-27)$ the needed derivatives of the weights can be expressed in terms of the corresponding derivatives of the $c_{ji}$ and the $a_{ji}$
\begin{align}
\frac{\partial w_{ji}}{\partial m_{jk}}&=\begin{cases}
\frac{\partial c_{j,i-1}}{\partial m_{jk}}x_{j,i-1}+\frac{\partial a_{ji}}{\partial m_{jk}}x_{ji} & k\neq i\\
1+\frac{\partial c_{j,i-1}}{\partial m_{ji}}x_{j,i-1}+\frac{\partial a_{ji}}{\partial m_{ji}}x_{ji} & k=i
\end{cases}\\
\frac{w_{ji}}{\partial \delta_j}&=\frac{\partial c_{j,i-1}}{\partial \delta_{j}}x_{j,i-1}+\frac{\partial a_{ji}}{\partial \delta_{j}}x_{ji}\\
\frac{w_{ji}}{\partial \gamma_j}&=\frac{\partial c_{j,i-1}}{\partial \gamma_{j}}x_{j,i-1}+\frac{\partial a_{ji}}{\partial \gamma_{j}}x_{ji}
\end{align}
where the index $j$ lies in the set $\{1,...,d\}$, while for a given $j$ the indices $i$ and $k$ lie in the set $\{1,...,K_j\}$. The derivatives of the $c_{ji}$ ($i=1,...,K_j-1$) with respect to the variational parameters are given by

\begin{small}
\begin{align}
\frac{\partial c_{ji}}{\partial m_{jk}}&=
\begin{cases}
(\rho_j\tau_j^2m_{ji}m_{j,i+1})u^{-\frac{3}{2}}c_{j,i-1}\frac{\partial c_{j,i-1}}{\partial m_{jk}} & k<i\\
\frac{v\left[\sqrt{u}-m_{ji}u^{-\frac{1}{2}}\left(\tau_j^2m_{ji}-c_{j,i-1}\frac{\partial c_{j,i-1}}{m_{ji}}\right)\right]}{u} & k=i\\
\frac{\rho_j\tau_j^2m_{ji}}{\sqrt{u}} & k=i+1\\
0 & k>i+1
\end{cases}\\
\frac{\partial c_{ji}}{\partial \delta_j}&=w_{\delta}\frac{vm_{ji}\left[2\sqrt{u}-\tau_ju^{-\frac{1}{2}}\left(\tau_jm_{ji}^2-c_{j,i-1}\frac{\partial c_{j,i-1}}{\partial \tau_j}\right)\right]}{\tau_ju}\\
\frac{\partial c_{ji}}{\partial \gamma_{j}}&=w_{\gamma}\frac{\tau_j^2m_{ji}m_{j,i+1}\left[\sqrt{u}-\rho_ju^{-\frac{1}{2}}(-c_{j,i-1}\frac{\partial c_{j,i-1}}{\partial \rho_j})\right]}{u}
\end{align}
\end{small}
where $u:=\tau_j^2m_{ji}^2-c_{j,i-1}^2$, $v:=\rho_j\tau_j^2m_{j,i+1}$, $w_{\delta}:=\operatorname{exp}(\delta_j)/[1+\operatorname{exp}(\delta_j)]$, and $w_{\gamma}:=\operatorname{exp}(-\gamma_j)/[1+\operatorname{exp}(-\gamma_j)]^2$, and obviously each derivative of $c_{j0}$ equals zero. Moreover, the derivatives of the $a_{ji}$ ($i=1,...,K_j-1$) with respect to the variational parameters are given by:
\begin{small}
\begin{align}
\frac{\partial a_{ji}}{\partial m_{jk}}&=
\begin{cases}
(\rho_j\tau_j^2|m_{ji}||m_{j,i+1}|)(-1)c_{ji}^{-2}\frac{\partial c_{ji}}{\partial m_{jk}} & k<i\\
\frac{\rho_j\tau_j^2|m_{j,i+1}|\left[\operatorname{sign}(m_{ji})c_{ji}-|m_{ji}|\frac{\partial c_{ji}}{\partial m_{ji}}\right]}{c_{ji}^2} & k =i\\
\frac{\rho_j\tau_j^2|m_{j,i}|\left[\operatorname{sign}(m_{j,i+1})c_{ji}-|m_{j,i+1}|\frac{\partial c_{ji}}{\partial m_{j,i+1}}\right]}{c_{ji}^2} & k=i+1\\
0 & k>i+1
\end{cases}\\
\frac{\partial a_{ji}}{\partial \delta_{j}}&=w_{\delta}\frac{\rho_j\tau_j|m_{ji}||m_{j,i+1}|\left[2c_{ji}-\tau_j\frac{\partial c_{ji}}{\partial \tau_j}\right]}{c_{ji}^2}\\
\frac{\partial a_{ji}}{\partial \gamma_{j}}&=w_{\gamma}\frac{\tau_j^2|m_{ji}||m_{j,i+1}|\left[c_{ji}-\rho_j\frac{\partial c_{ji}}{\partial \rho_j}\right]}{c_{ji}^2}
\end{align}
\end{small}
In addition, the derivatives of $a_{jK_j}$ with respect to the variational parameters are given by
\begin{small}
\begin{align}
\frac{\partial a_{jK_j}}{\partial m_{jk}}&=
\begin{cases}
(-1)y^{-\frac{1}{2}}c_{j,K_{j}-1}\frac{\partial c_{j,K_{j}-1}}{\partial m_{jk}} & k<K_j\\
y^{-\frac{1}{2}}\left(\tau_j^2m_{jK_j}-c_{j,K_{j}-1}\frac{\partial c_{j,K_{j}-1}}{\partial m_{jK_j}}\right) & k=K_j
\end{cases}\\
\frac{\partial a_{jK_j}}{\partial \delta_{j}}&=w_{\delta}y^{-\frac{1}{2}}\left(\tau_jm_{jK_j}^2-c_{j,K_j-1}\frac{\partial c_{j,K_{j}-1}}{\tau_j}\right)\\
\frac{\partial a_{jK_j}}{\partial \gamma_{j}}&=w_{\gamma}y^{-\frac{1}{2}}\left(-c_{j,K_{j}-1}\frac{\partial c_{j,K_{j}-1}}{\rho_j}\right)
\end{align}
\end{small}
where $y:=\tau_j^2m_{jK_j}^2-c_{j,K_{j}-1}^2$.

Finally, the partial derivatives of the KL divergence $D_{KL}(q_{\boldsymbol{\phi}}(\mathbf{w})||p(\mathbf{w}))$ (abbreviated with $D_{KL}$) with respect to the variational parameters are given by:
\begin{small}
\begin{align}
\frac{\partial}{\partial m_{jk}}D_{KL}&=-\sum\limits_{i=1}^{K_j}\left(\frac{1}{a_{ji}}\frac{\partial a_{ji}}{\partial m_{jk}}\right)+\frac{\tau_j^2}{\zeta_j^2}m_{jk}+\frac{1}{\zeta_j^2}(m_{jk}-\mu_{jk})\\
\frac{\partial}{\partial \delta_{j}}D_{KL}&=-\sum\limits_{i=1}^{K_j}\left(\frac{1}{a_{ji}}\frac{\partial a_{ji}}{\partial \delta_{j}}\right)+\frac{\operatorname{exp}(\delta_j)}{1+\operatorname{exp}(\delta_j)}\frac{\tau_j}{\zeta_j^2}||\mathbf{m}_j||_2^2\\
\frac{\partial}{\partial \gamma_{j}}D_{KL}&=-\sum\limits_{i=1}^{K_j}\left(\frac{1}{a_{ji}}\frac{\partial a_{ji}}{\partial \gamma_{j}}\right)
\end{align}
\end{small}
For reasons of brevity the calculation process corresponding to the derivatives presented is not reported.

\subsection{Implementation and Pseudocode}
\label{pseudocode}

We have implemented the proposed apporach by modifying and extending the popular open-source deep learning framework Caffe, see \cite{jia2014caffe}. In particular, our implementation includes a Bayesian version of the classical inner product layer as well as the convolutional layer. This allows for the training of (deep) multilayer perceptrons (MLPs) and, moreover, convolutional neural networks (CNNs) according to our novel approach. Up to now we have not parallelized our code such that it can run on GPU. This is left for future research and will enable the training of state-of-the-art CNNs for image classification in a reasonable amount of time. The pseudocode which was starting point of our implementation is presented below. The code shows how a classical, i.e. frequentist, inner product, or convolutional layer can be extended in order to fit with the methodology presented.

Besides parameters, for which obviously initial values are required, the presented pseudo code also asks for the two additional parameters $\nu$ and $\kappa$. Empirically, we discovered that in practical applications it can be helpful to use another penalization strength of the KL divergence than the fixed one proposed in Section \ref{background}. In particular, the derivatives of the KL divergence do not suffer from the vanishing gradients problem in contrast to the derivatives of the loss function. For this reason, reducing the penalization strength of the divergence by a fixed factor, which is equivalent to reducing the learning rate of the divergence, might be a good decision. Moreover, again, empirically we discovered that the derivatives with respect to the $\gamma_j$ are quite small which slows down the learning process. To overcome this problem the  learning rate multiplier $\kappa$ can be used.

In the forward pass at first variational parameters which hold 	``critical'' values are assigned similar but less ``critical'' ones. This way of proceeding should guarantee that no numerical issues occur during the training process. Thus, very small expectation values, variances (smaller than $0.01$ times abs(corresponding expectation value) in absolute values), and correlations (smaller than $0.01$ in absolute values) are replaced, as well as correlations which are nearly given by $0.5$, or $-0.5$. Recall that the correlations have to stay in the interval $(-\frac{1}{2},\frac{1}{2})$ in order for the covariance matrices to be positive definite. After the checkings for numerical stability the weights and the biases of the network are sampled from the variational distribution. The sampled parameters are then used in place of the deterministic ones in non-Bayesian deep learning in order to perform the classical forward pass.

In the backward pass at first the derivatives of the loss function with respect to the sampled weights and biases are computed. This is done completely analogous as in classical deep learning, but the sampled parameters are used in place of the deterministic ones. In the next steps the derivatives of the sampled network parameters with respect to the variational parameters are computed, and the derivatives of the KL divergence with respect to the variational parameters are calculated. Finally, an appropriate merging of all the computed derivatives results in the derivatives of the approximation $L_{VI}^k$ of the negative log evidence lower bound. These derivatives are used to update the variational parameters according to some learning schedule.

\begin{algorithm}
\caption{Bayes deep learning layer}
\begin{algorithmic}[1]
\REQUIRE ~\\
\begin{enumerate}[-]
\setlength{\itemsep}{0pt}
\item Initial variational parameters $\mathbf{m}_j,\delta_j,\gamma_j,\mathbf{b}_{j},\delta_{bj}$ and $\gamma_{bj}$
\item Factor $\nu$ for penalization strength of $D_{KL}$
\item Learning rate multiplier $\kappa$ for $\gamma_j$
\item Parameters $\boldsymbol\mu_j,\zeta_j,\boldsymbol\mu_{bj}$ and $\zeta_{bj}$ for the prior
\item Number of training iterations $N_{iter}$
\end{enumerate}

\FOR{$i \text{ in } 1:N_{iter}$}
\STATE
\STATE \textbf{Forward pass:}
\STATE
\STATE \textit{Guarantee numerical stability:}
\IF{$\gamma_j\in (-0.04000533,0.04000533)$}
\STATE Set $\gamma_j$ to $0.04000533$ with probability $\frac{1}{2}$ and to\\ \hspace*{5mm} $-0.04000533$ otherwise
\ENDIF
\IF{$\gamma_j > 10 $}
\STATE  Set $\gamma_j=10$ 
\ENDIF
\IF{$\gamma_j < -10 $}
\STATE  Set $\gamma_j=-10$ 
\ENDIF
\IF{$\delta_j < -4.600166 $}
\STATE  Set $\delta_j=-4.600166$ 
\ENDIF
\FOR{$k \text{ in } 1:K_j$}
\IF{$m_{jk} \in (-0.000001,0.000001) $}
\STATE  Set $m_{jk}$ to $0.000001$ with probability $\frac{1}{2}$ and to\\ \hspace*{10mm} $-0.000001$ otherwise
\ENDIF
\ENDFOR
\STATE
\STATE \textit{Sample from the variational distribution:}
\STATE Draw $K_j$ independent samples from $\mathcal{N}(0,1)$ and

\hspace*{-3.6mm}thus a sample $\mathbf{x}_j$ from $\mathcal{N}(\mathbf{0}_{K_j},\mathbf{I}_{K_j})$
\STATE Compute $\mathbf{L}_j$ according to Equations (17-23)
\STATE Set $\mathbf{w}_j=\mathbf{m}_j+\mathbf{L}_j\mathbf{x}_j$
\STATE
\STATE Repeat lines $(6-30)$, but now for the biases
\STATE Use the the sampled weights and biases in place of

\hspace*{-3.6mm}the classical weights and biases in non-Bayesian

\hspace*{-3.6mm}deep learning and proceed as in the classical case
\STATE
\STATE \textbf{Backward pass}
\STATE
\STATE Treat the sampled weights as the classical ones in

\hspace*{-3.6mm}non-Bayesian deep learning in order to compute the

\hspace*{-3.6mm}derivatives of the loss function $L$ with respect to

\hspace*{-3.6mm}them
\algstore{myalg}
\end{algorithmic}
\end{algorithm}

\begin{algorithm}                     
\begin{algorithmic} [1]                   
\algrestore{myalg}
\STATE \textit{Compute the derivatives of $L_{VI}^k$ with respect to $\mathbf{m}_j$}:
\STATE Define: $a_{Deriv} = 0,c_{NewDeriv} = 0,$ $c_{OldDeriv} = 0$
\STATE Declare a variable $wDeriv$
\STATE Declare a $K_j$ dimensional array $\text{diff}_{m_j}$, which will

\hspace*{-3.6mm}hold the derivative $\frac{\partial L_{VI}^k}{\partial m_{jk}}$ as $k$-th entry
\STATE Initialize $\text{diff}_{m_j}$ with zeros
\FOR{$k \text{ in } 1:K_j$}
\FOR{$l \text{ in } 1:K_j-1$}
\STATE Assign the derivative $\frac{\partial c_{jl}}{\partial m_{jk}}$ according to

\hspace*{3.6mm}Equation $(37)$ to $c_{NewDeriv}$ (note: $c_{OldDeriv}$ \hspace*{10.6mm}holds $\frac{\partial c_{j,l-1}}{\partial m_{jk}}$)
\STATE Assign the derivative $\frac{\partial a_{jl}}{\partial m_{jk}}$ according to

\hspace*{3.6mm}Equation $(40)$ to $a_{Deriv}$
\STATE Assign the derivative $\frac{\partial w_{jl}}{\partial m_{jk}}$ according to

\hspace*{3.6mm}Equation $(34)$ to $w_{Deriv}$
\STATE According to Equation $(31)$ add $\frac{\partial L}{\partial  w_{jl}}w_{Deriv}$ to

\hspace*{3.6mm}$\text{diff}_{m_j}[k]$
\STATE According to Equation $(46)$ subtract $\nu \frac{a_{Deriv}}{a_{jl}}$

\hspace*{3.6mm}from $\text{diff}_{m_j}[k]$
\STATE Set $c_{OldDeriv}=c_{NewDeriv}$
\ENDFOR
\STATE Set $l=K_j$
\STATE Assign the derivative $\frac{\partial a_{jl}}{\partial m_{jk}}$ according to

Equation $(43)$ to $a_{Deriv}$
\STATE Assign the derivative $\frac{\partial w_{jl}}{\partial m_{jk}}$ according to

Equation $(34)$ to $w_{Deriv}$
\STATE According to Equation $(31)$ add $\frac{\partial L}{\partial  w_{jl}}w_{Deriv}$ to

$\text{diff}_{m_j}[k]$
\STATE According to Equation $(46)$ subtract $\nu \frac{a_{Deriv}}{a_{jl}}$

from $\text{diff}_{m_j}[k]$ and add the term

$\nu\left[\frac{\tau_j^2}{\zeta_j^2}m_{jk}+\frac{1}{\zeta_j^2}(m_{jk}-\mu_{jk})\right]$
\ENDFOR
\STATE
\STATE \textit{Compute the derivatives of $L_{VI}^k$ with respect to $\delta_j$}:
\STATE Compute the derivatives $\frac{\partial c_{ji}}{\partial \delta_j}$ $(i=1,...,K_j-1)$ 

\hspace*{-3.6mm}according to Equation $(38)$ 
\STATE Compute the derivatives $\frac{\partial a_{ji}}{\partial \delta_j}$ $(i=1,...,K_j)$ 

\hspace*{-3.6mm}according to the Equations $(41)$ and $(44)$
\STATE Compute the derivatives $\frac{\partial w_{ji}}{\partial \delta_j}$ $(i=1,...,K_j)$ 

\hspace*{-3.6mm}according to Equation $(35)$
\STATE Compute the derivative $\frac{\partial L}{\partial \delta_j}$ according to Equation

\hspace*{-3.6mm}$(32)$
\STATE In order to finally obtain $\frac{\partial L_{VI}^k}{\partial \delta_j}$ add $\nu\frac{\partial}{\partial \delta_j}D_{KL}$

\hspace*{-3.6mm}according to Equation $(47)$ to $\frac{\partial L}{\partial \delta_j}$
\algstore{myalg2}
\end{algorithmic}
\end{algorithm}

\begin{algorithm}                     
\begin{algorithmic} [1]                   
\algrestore{myalg2}
\STATE \textit{Compute the derivatives of $L_{VI}^k$ with respect to $\gamma_j$}:
\STATE Compute the derivatives $\frac{\partial c_{ji}}{\partial \gamma_j}$ $(i=1,...,K_j-1)$ 

\hspace*{-3.6mm}according to Equation $(39)$ 
\STATE Compute the derivatives $\frac{\partial a_{ji}}{\partial \gamma_j}$ $(i=1,...,K_j)$ 

\hspace*{-3.6mm}according to the Equations $(42)$ and $(45)$
\STATE Compute the derivatives $\frac{\partial w_{ji}}{\partial \gamma_j}$ $(i=1,...,K_j)$ 

\hspace*{-3.6mm}according to Equation $(36)$
\STATE Compute the derivative $\frac{\partial L}{\partial \gamma_j}$ according to Equation

\hspace*{-3.6mm}$(33)$
\STATE In order to finally obtain $\frac{\partial L_{VI}^k}{\partial \gamma_j}$ add $\nu\frac{\partial}{\partial \gamma_j}D_{KL}$

\hspace*{-3.6mm}according to Equation $(48)$ to $\frac{\partial L}{\partial \gamma_j}$ 
\STATE Multiply $\frac{\partial L_{VI}^k}{\partial \gamma_j}$ with $\kappa$
\STATE
\STATE Compute the gradient with respect to bottom data

\hspace*{-3.6mm}as it is done in frequentist deep learning, but use

\hspace*{-3.6mm}the sampled weights $w_j$ in place of the deterministic

\hspace*{-3.6mm}ones
\STATE Repeat lines $(38-73)$, but now for the biases
\STATE Use the computed derivatives of $L_{VI}^k$ with respect

\hspace*{-3.6mm}to the variational parameters in order to update

\hspace*{-3.6mm}them according to some chosen learning schedule
\STATE
\ENDFOR
\end{algorithmic}
\end{algorithm}

\section{Experiments}
\label{experiments}
This section investigates how well the proposed approach performs on real world datasets. On the one hand the prediction accuracy and on the other hand the quality of the uncertainty information provided are of particular interest. The popular benchmark datasets MNIST \cite{Deng2012} and CIFAR-10 \cite{Krizhevsky2009LearningML} form basis of the evaluations.

\subsection{MNIST}
\label{mnist}
In this section, the performance of the proposed method is evaluated based on the MNIST dataset. This dataset contains $70,000$ grayscale images of handwritten digits ($0-9$). The complete dataset is partitioned in a training dataset which counts $60,000$ images and a test set of $10,0000$ examples. Each image is of size $28\times 28$.

The architecture chosen for the performance evaluation is the popular LeNet proposed by Lecun et al. \cite{Lecun1998}. This CNN mainly consists of two convolutional layers and, further, two fully connected layers. The special version of LeNet used in this paper has the following additional specifications: The ReLU activation function is assigned to the first fully connected layer, while the other layers simply use the identity function as activation function. Moreover, the first convolutional layer includes $20$ and the second one includes $50$ $5\times 5$ kernels. Max-pooling with kernel size $2\times 2$ and stride of $2$ is applied after both convolutional layers. The number of neurons of the second fully connected layer is determined by the number of possible classes and thus given by $10$. In contrast to the second fully connected layer the number of neurons in the first fully connected layer can be freely specified. In our experiments we set this number once to $100$ and once to $250$.

To get an idea how well the proposed Bayesian approach predicts compared to the frequentist one, we train LeNet according to both approaches. Mini-batch gradient descent with a batch size of $64$ is chosen for optimization. Further, a decreasing learning rate policy is selected. In particular, the learning rate in the $k$-th iteration is specified as $0.01\cdot(1 + 0.0001 \cdot i)^{−0.75}$. Moreover, momentum is applied with a value of $0.9$. The total amount of iterations is set to $100,000$. In terms of regularization in the frequentist approach a penalization of the Euclidean norm with a penalization strength of $0.0005$ takes place. Further, dropout is applied after the first fully connected layer with a dropping rate given by $0.5$. These regularization techniques are not applied in the Bayesian approach which is naturally regularized due to the sampling from the variational distribution and the penalization of the KL divergence during network training. However, in Bayesian deep learning the parameters of the prior have to be specified. We set the expectation values of all network parameters to $0$ and the variances to $1$. There is no a priori knowledge available such that the prior can merely be used to guarantee that network parameters do not diverge. Moreover, for each network layer we set the parameters $\nu$ and $\kappa$ described in Section \ref{pseudocode} to $1/(60,000\cdot 100)$ and $50$, respectively. Thus, the penalization strength of the KL divergence is reduced by a factor of $100$ due to the reasons described in Section \ref{pseudocode}.

\begin{figure*}
    \centering
    \begin{minipage}[t]{0.45\linewidth}
\centering
\includegraphics[width=0.9\textwidth]{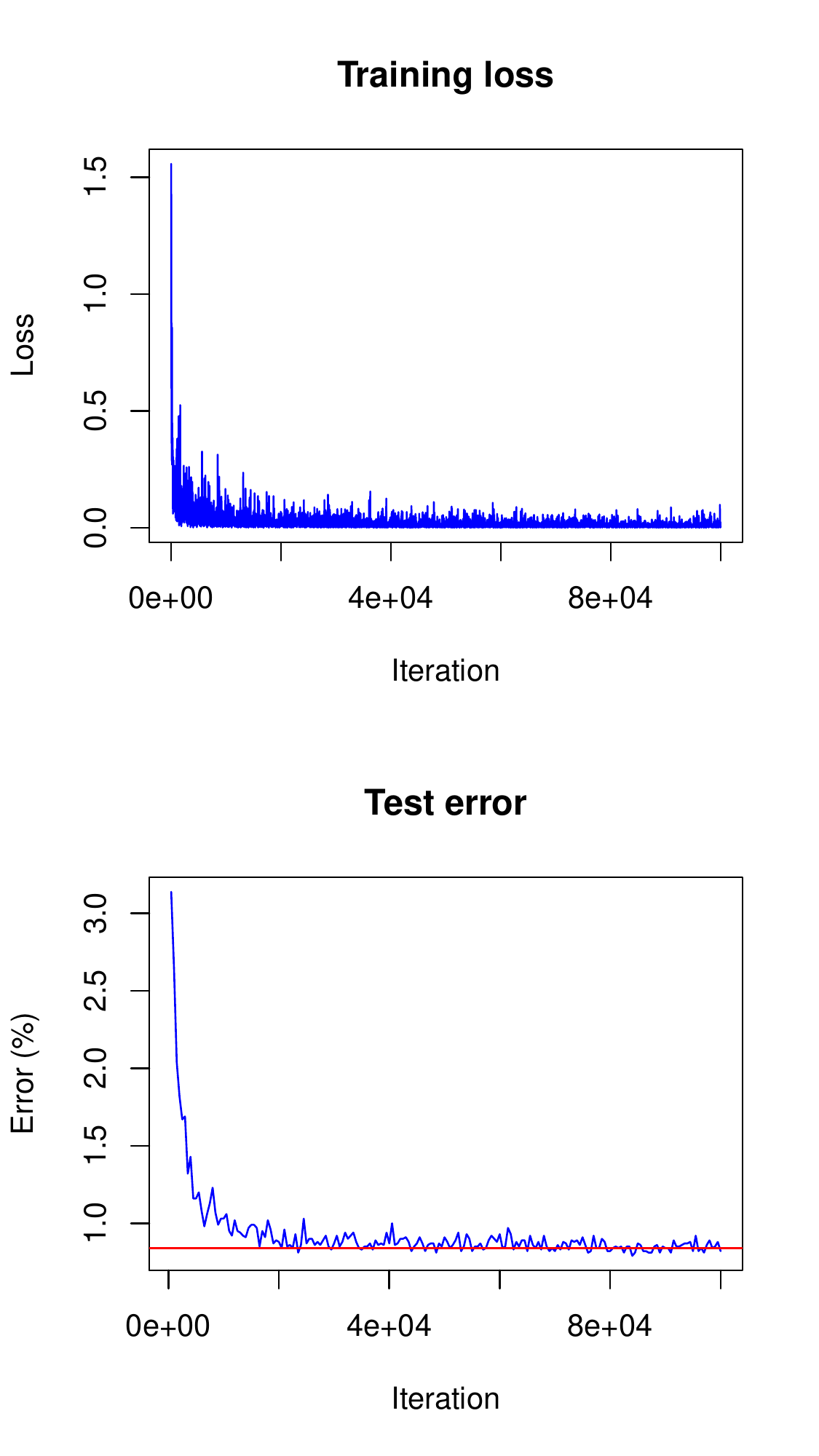}
\caption{Training visualization of frequentist LeNet with
$100$ neurons in the first fully connected layer. The horizontal line marks the achieved test error.}  
\label{freqmnist} 
    \end{minipage}
    \hspace{0.5cm}
    \begin{minipage}[t]{0.45\linewidth}
    \centering
\includegraphics[width=0.9\textwidth]{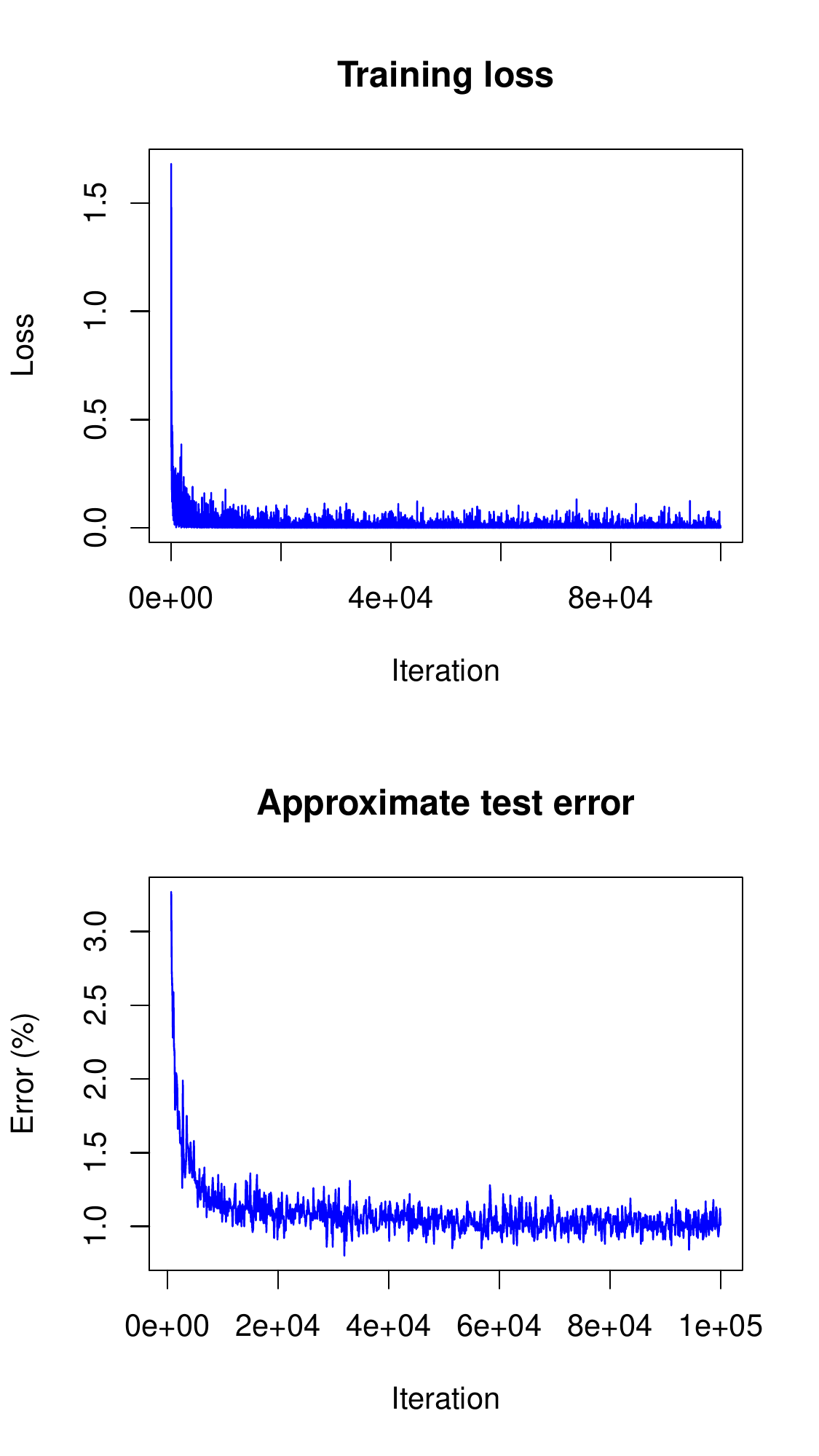}
\caption{Training visualization of Bayesian LeNet with
$100$ neurons in the first fully connected layer. For the computation of the approximate test error only one sample is drawn from the variational distribution per image.}  
\label{bayesmnist} 
    \end{minipage}  
\end{figure*}

For the model with $100$ neurons in the first fully connected layer both the Bayesian and the frequentist training process are visualized in the Figures \ref{freqmnist} and \ref{bayesmnist}. One can see that the loss decreases in a quite similar way in both processes. This is an interesting result since in the work of Posch et al. \cite{Posch2019} the Bayesian loss fluctuates heavily compared to the classical one. Thus, the learning of correlations between network parameters (which is our extension of their work) enables a smoother network training. Note that the test error plotted in Figure \ref{bayesmnist} is just a rough approximation of the true one, which is based on one sample from the variational distribution per test image. Usually, the corresponding true errors are significantly lower. However, the rough approximation suffices to monitor network training. A more accurate and thus also computationally more expensive approximation based on multiple samples can be made at the end of the training process.

The finally achieved test errors are given in Table \ref{table:errormnist}. Note that the predictions of the Bayesian models are based on $200$ samples from the corresponding variational distributions per test image. One can see that the Bayesian models perform a little bit better than their frequentist analogues. The reason for the superior results of the Bayesian models can be found in their natural robustness against overfitting. It can be hard to regularize frequentist models in such a way that they obtain a robustness comparable to Bayesian models.
\begin{table}
\caption{Test errors of the trained models. The predictions of the Bayesian models are based on $200$ samples from the corresponding variational distributions per test image.}
\label{table:errormnist}
\centering 
\begin{tabularx}{\linewidth}{ l@{\hskip 1in}  l }
  \hline			
  Model & Test error  \\
  \hline\\
  Frequentist (100 neurons) & 0.84\%  \\
  Bayesian (100 neurons) & 0.7\% \\
  Frequentist (250 neurons) & 0.7\% \\
  Bayesian (250 neurons) &0.61\%  \\
  \hline  
\end{tabularx}
\end{table}

In Table \ref{table:variationalmnist} the correlations $\rho_j$ and the variance determining parameters $\tau_j$ of the two Bayesian models can be found ($j=1,...,4$). One can see that there is a low a posteriori uncertainty about the bias terms and further that they are nearly uncorrelated. This is a plausible result since the number of bias terms is small compared to the number of weights. The a posteriori uncertainty of the weights differs significantly from layer to layer. While the first convolutional layer and the second fully connected layer go along with a low a posteriori uncertainty about the network weights, the other layers show a high uncertainty. For the network with $250$ neurons in the first fully connected layer the standard deviation of the posterior distribution of a weight is given by two times the corresponding expectation value. Note that layers with low uncertainty are exactly the ones which directly act on the network input and the network output. The correlations of the weights are all negative.
\begin{table}
\caption{Parameters $\rho_j$ and $\tau_j$ corresponding to  the variational distributions of the Bayesian models with $100$ and $250$ neurons in the first fully connected layer ($j=1,...,4$).}
\label{table:variationalmnist}
\centering 
\small
\begin{tabularx}{\linewidth}{ l l l l l l }
  \hline		
  Layer & Neurons &  $\tau_j$ &  $\tau_{bj}$ &  $\rho_{j}$ &  $\rho_{bj}$ \\
  \hline\\
  Convolutional 1   & 100 & 0.04 & 0.05 & -0.44 & -0.04  \\
  Convolutional 2   & 100 & 0.52 & 0.05 & -0.25 & -0.01  \\
  Fully connected 1 & 100 & 0.86 & 0.05 & -0.15 & ~0.01\\
  Fully connected 2 & 100 & 0.05 & 0.04 & -0.21 & ~0.01\\
  Convolutional 1   & 200 & 0.03 & 0.05 & -0.44 & ~0.03  \\
  Convolutional 2   & 200 & 0.35 & 0.05 & -0.21 & -0.01  \\
  Fully connected 1 & 200 & 2.02 & 0.06 & -0.15 & -0.01\\
  Fully connected 2 & 200 & 0.06 & 0.05 & -0.18 & ~0.01\\
  \hline  
\end{tabularx}
\end{table}

\begin{figure}
\centering
\includegraphics[width=0.95\linewidth]{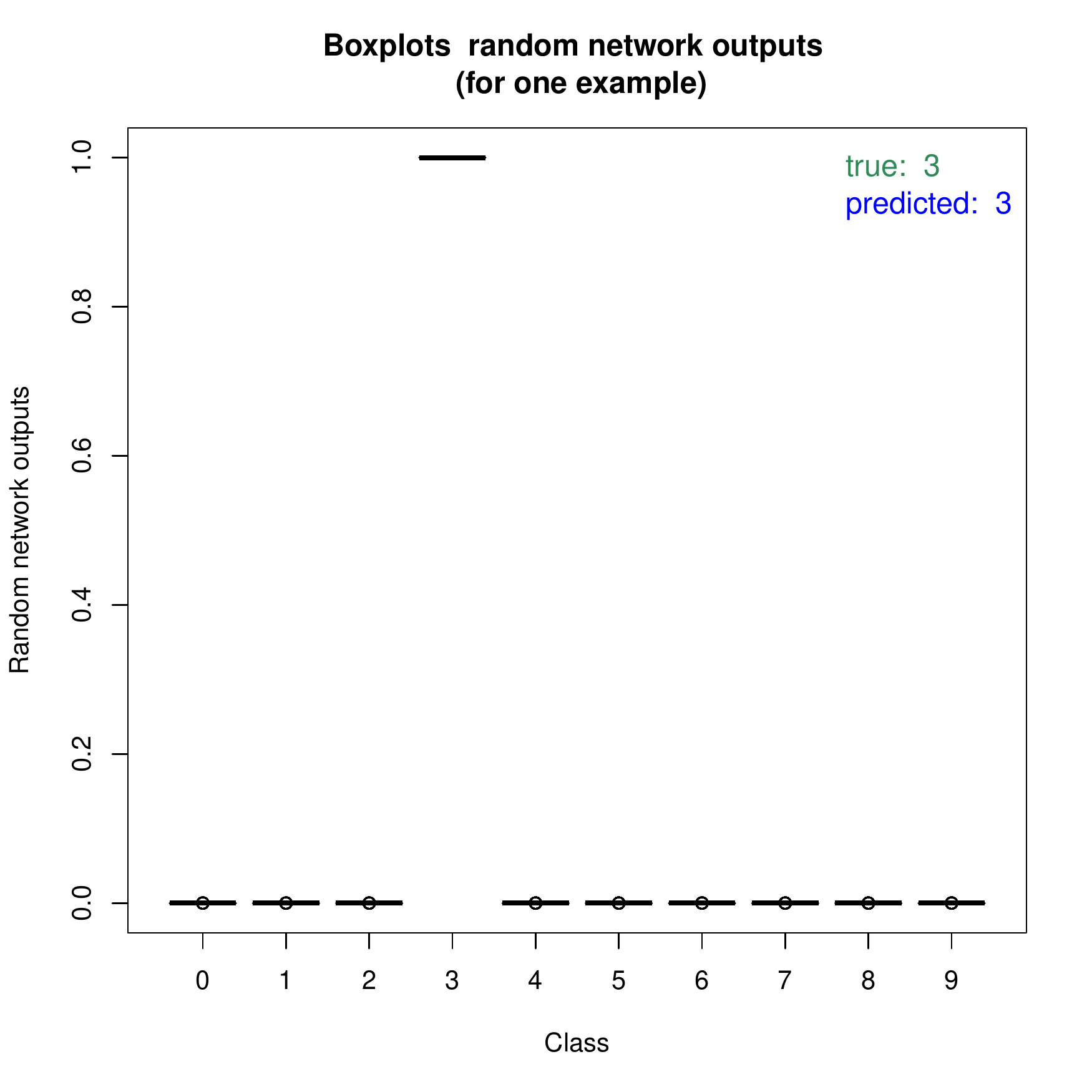}
\caption{Boxplots of $200$ random network outputs for a representative correct classification result. Model with $100$ neurons in the first fully connected layer.}  
\label{boxmnistcorr} 
\end{figure}

\begin{figure}
\centering
\includegraphics[width=0.95\linewidth]{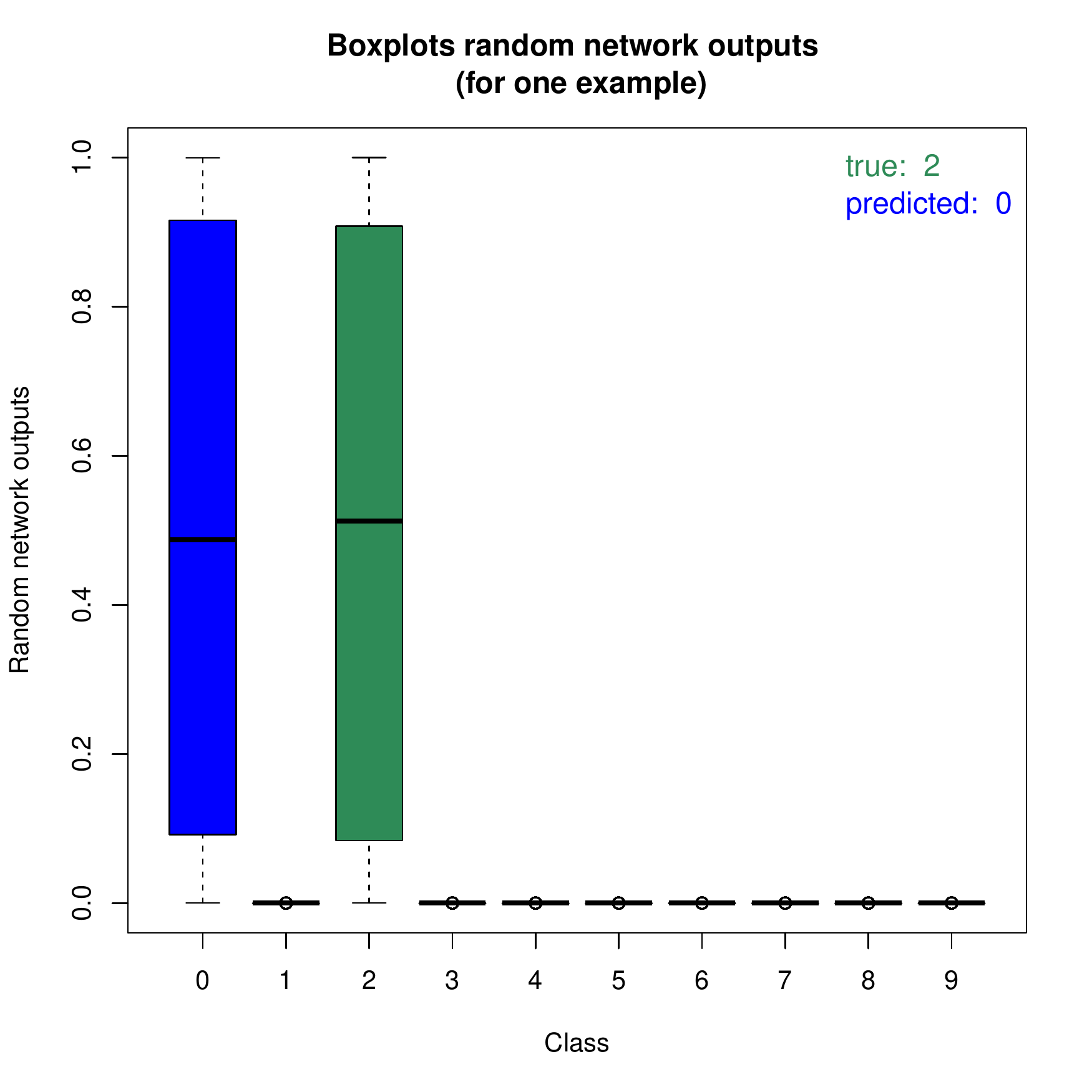}
\caption{Boxplots of $200$ random network outputs for a representative incorrect classification result. Model with $100$ neurons in the first fully connected layer.}  
\label{boxmnistfalse1} 
\end{figure}

\begin{figure}
\centering
\includegraphics[width=0.95\linewidth]{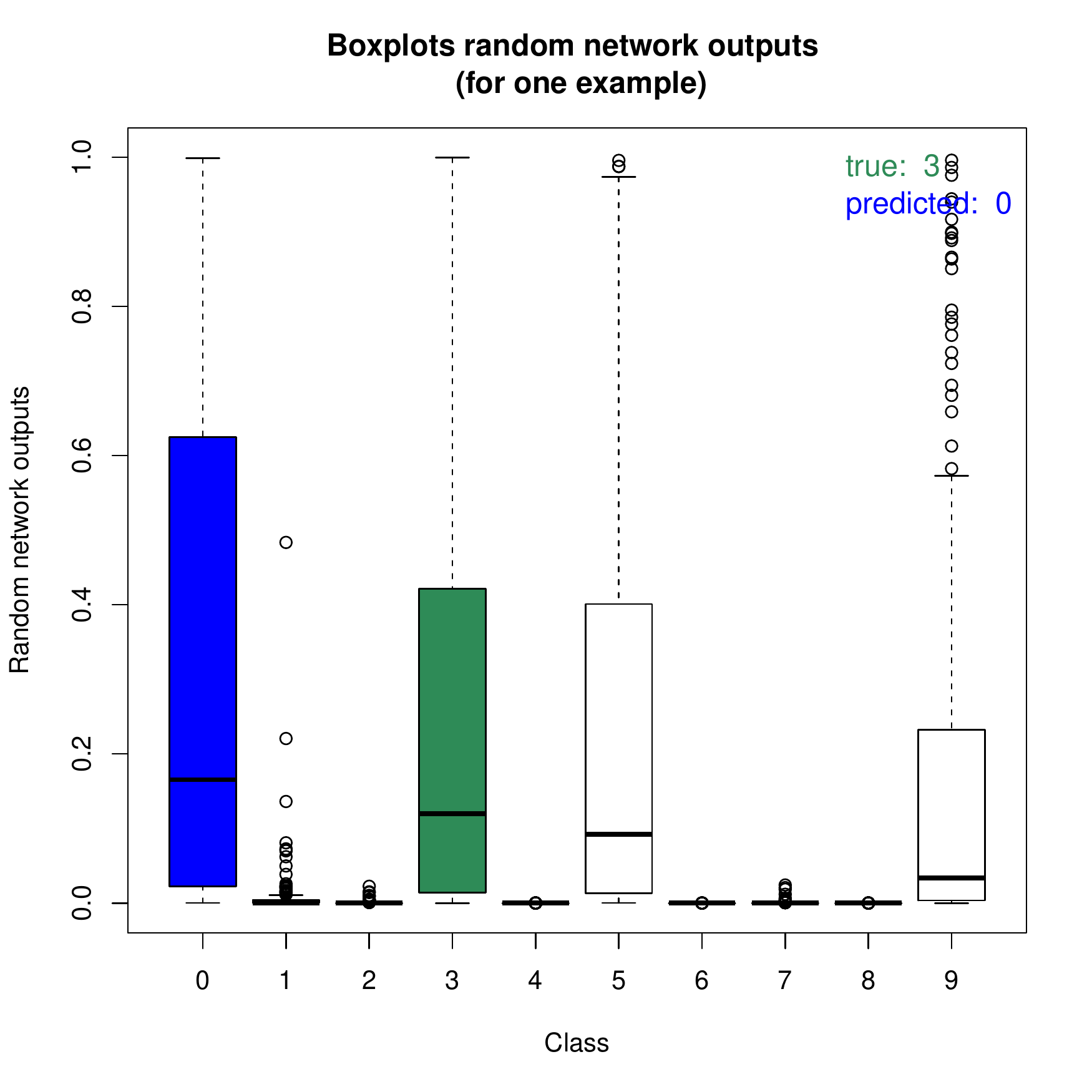}
\caption{Boxplots of $200$ random network outputs for a representative incorrect classification result. Model with $100$ neurons in the first fully connected layer.}  
\label{boxmnistfalse2} 
\end{figure}

\begin{figure*}
\centering
    \begin{subfigure}{.1\textwidth}
        \centering
        \includegraphics[width=\textwidth]{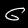}
        \caption{True: 6\\
        Predicted: 5}
    \end{subfigure}%
    ~
    \begin{subfigure}{.1\textwidth}
        \centering
        \includegraphics[width=\textwidth]{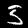}
        \caption{True: 5\\
        Predicted: 3}
    \end{subfigure}%
    ~
    \begin{subfigure}{0.1\textwidth}
        \centering
        \includegraphics[width=\textwidth]{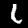}
        \caption{True: 6\\
        Predicted: 1}
    \end{subfigure}%
    ~
    \begin{subfigure}{0.1\textwidth}
        \centering
        \includegraphics[width=\textwidth]{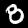}
        \caption{True: 8\\
        Predicted: 0}
    \end{subfigure}%
    ~
    \begin{subfigure}{0.1\textwidth}
        \centering
        \includegraphics[width=\textwidth]{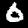}
        \caption{True: 6\\
        Predicted: 0}
    \end{subfigure}%
    ~
    \begin{subfigure}{0.1\textwidth}
        \centering
        \includegraphics[width=\textwidth]{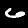}
        \caption{True: 6\\
        Predicted: 4}
    \end{subfigure}%
    ~
    \begin{subfigure}{0.1\textwidth}
        \centering
        \includegraphics[width=\textwidth]{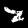}
        \caption{True: 2\\
        Predicted: 7}
    \end{subfigure}%
    ~
    \begin{subfigure}{0.1\textwidth}
        \centering
        \includegraphics[width=\textwidth]{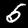}
        \caption{True: 5\\
        Predicted: 6}
    \end{subfigure}
\caption{All test images where the model with $250$ neurons in the first fully connected layer was quite certain in its wrong prediction result.}
\label{mnistimages}
\end{figure*}

As already mentioned in Section \ref{background} the uncertainties in the network parameters of Bayesian nets can directly be translated in uncertainty about the predictions. In particular, credible intervals for the probability that an image shows an object of a given class can be estimated by computing multiple neural network outputs  with weights sampled from the variational distribution. Figures \ref{boxmnistcorr}, \ref{boxmnistfalse1}, and \ref{boxmnistfalse2} show boxplots of respectively $200$ random network outputs corresponding to three representative images from the test dataset. The network which was used to produce these figures is the one with $100$ neurons in the first fully connected layer. In Figure \ref{boxmnistcorr} there are no boxes at all since the network is very sure about its correct prediction. This is the case for most of the correctly classified images. The Figures \ref{boxmnistfalse1} and \ref{boxmnistfalse2} reflect the common behavior of the net in case of incorrect classification results. Either the network has difficulties to decide between two classes where one of the two is the true one, or it is completely uncertain what class to predict. To quantify the quality of the prediction uncertainty of the Bayesian models we estimate $95\%$ credible intervals of the component-wise network outputs for all test images. The estimates are based on $200$ random network outputs, respectively. A prediction result is considered as quite certain if the credible interval of the predicted class does not overlap with the intervals of the other classes. In the other case a prediction result is considered as uncertain. Table \ref{table:preduncmnist} summarizes the corresponding results.  One can see that the models are most of the times quite certain about correct prediction results while they are usually uncertain in wrong prediction results. This shows that the uncertainty information obtained from the proposed Bayesian approach is very good. In Figure \ref{mnistimages} all the test images can be found where the model with $250$ neurons in the first fully connected layer was quite certain about its wrong prediction. 
\begin{table}
\caption{Overview of quite certain and uncertain prediction results.}
\label{table:preduncmnist}
\centering 
\small
\begin{tabularx}{\linewidth}{ l l l l  }
  \hline		
  Model & Prediction result &  Quite certain &  Uncertain \\
  \hline\\
  100 neurons   & correct & 9668 & 262  \\
  100 neurons   & wrong   & 14 & 56   \\
  250 neurons   & correct & 9587 & 352   \\
  250 neurons   & wrong   & 8 & 53   \\
  \hline  
\end{tabularx}
\end{table}

\subsection{CIFAR-10}
In this section, the performance of the proposed method is evaluated based on the CIFAR-10 dataset. This dataset consists of $60,000$ RGB images in $10$ classes (airplane, automobile, bird, cat, deer, dog, frog, horse, ship, truck). The complete dataset is partitioned into a training dataset which counts $50,000$ images and a test set of $10,000$ examples. Each image is of size $32\times 32$.
The architecture chosen for the performance evaluation is the one included under the name \textit{CIFAR10\_full} in the Caffe framework \cite{jia2014caffe}. This CNN mainly consists of three convolutional layers followed by one fully connected layer. For further details take a look at the model definition available in the Caffe framework.

As in Section \ref{mnist} we train the network according to the frequentist approach and according to the proposed Bayesian approach. Mini-batch gradient descent with a batch size of $100$ is chosen for optimization. Further, a fixed learning rate policy with a learning rate of $0.001$ is selected. The total amount of training iterations is given by $100,000$ for the frequentist approach and by $40,000$ for the Bayesian approach. A penalization of the Euclidean norm with a penalization strength of $0.004$ for the convolutional layers and a penalization strength of $1$ for the fully connected layer is used for regularization of the frequentist net. In the Bayesian case the prior distribution is used for regularization. Therefore, we assign independent normal distributions with zero mean (see Section \ref{variationaldist}) to the network parameters. The standard deviations of these distributions are specified as $1$ for the convolutional layers and specified as $0.05$ for the fully connected layer. Thus, the fully connected layer gets stronger regularized than the other ones, as in the frequentist case. Moreover, for each network layer we set the parameters $\nu$ and $\kappa$ described in Section \ref{pseudocode} to $1/(50,000\cdot 10)$ and $50$. Thus, the penalization strength of the KL divergence is reduced by a factor of $10$ due to the reasons described in Section \ref{pseudocode}.

\begin{figure*}
    \centering
    \begin{minipage}[t]{0.45\linewidth}
\centering
\includegraphics[width=0.9\textwidth]{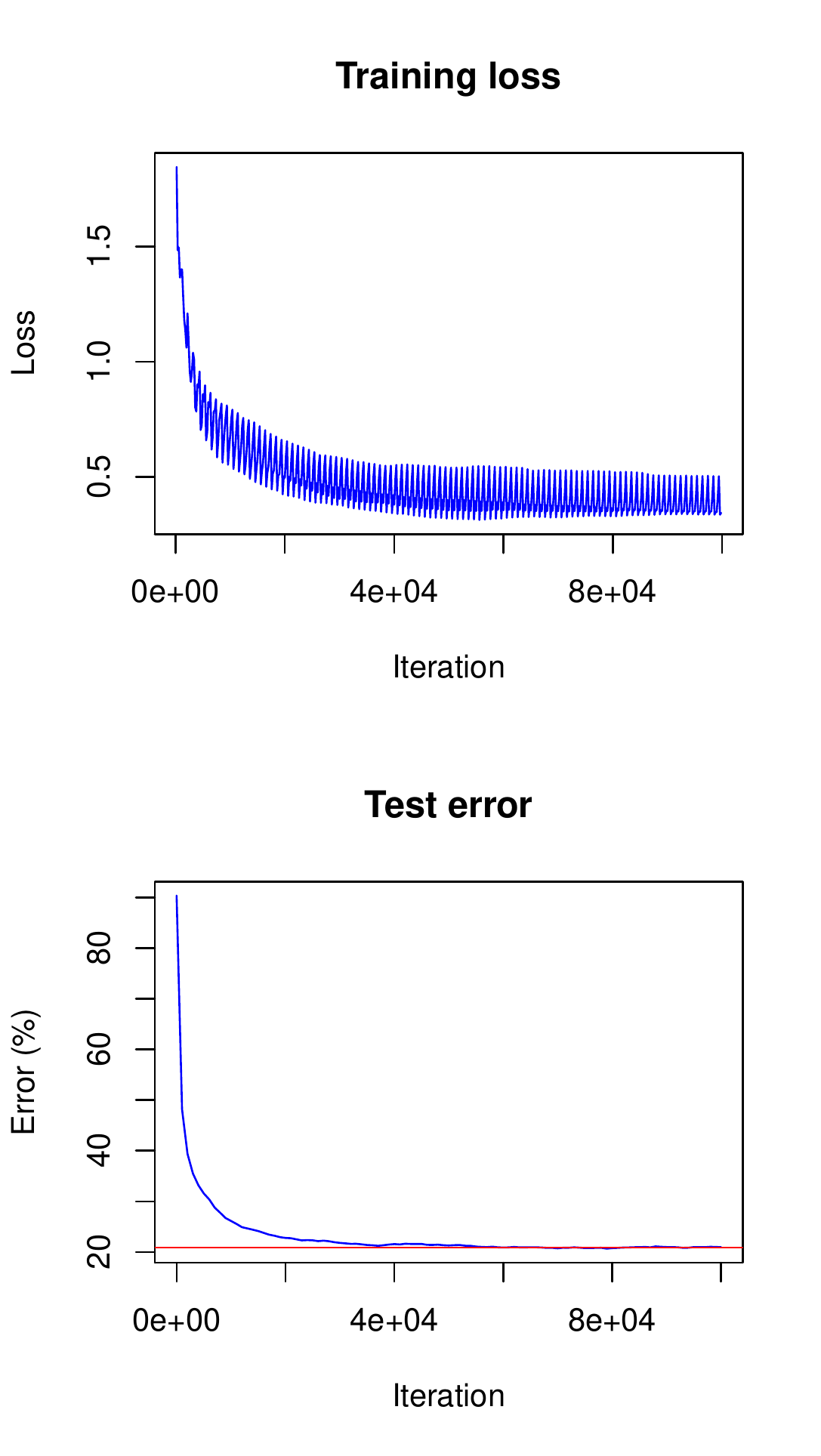}
\caption{Training visualization of the frequentist model. The horizontal line marks the achieved test error.}  
\label{freqcifar10} 
    \end{minipage}
    \hspace{0.5cm}
    \begin{minipage}[t]{0.45\linewidth}
    \centering
\includegraphics[width=0.9\textwidth]{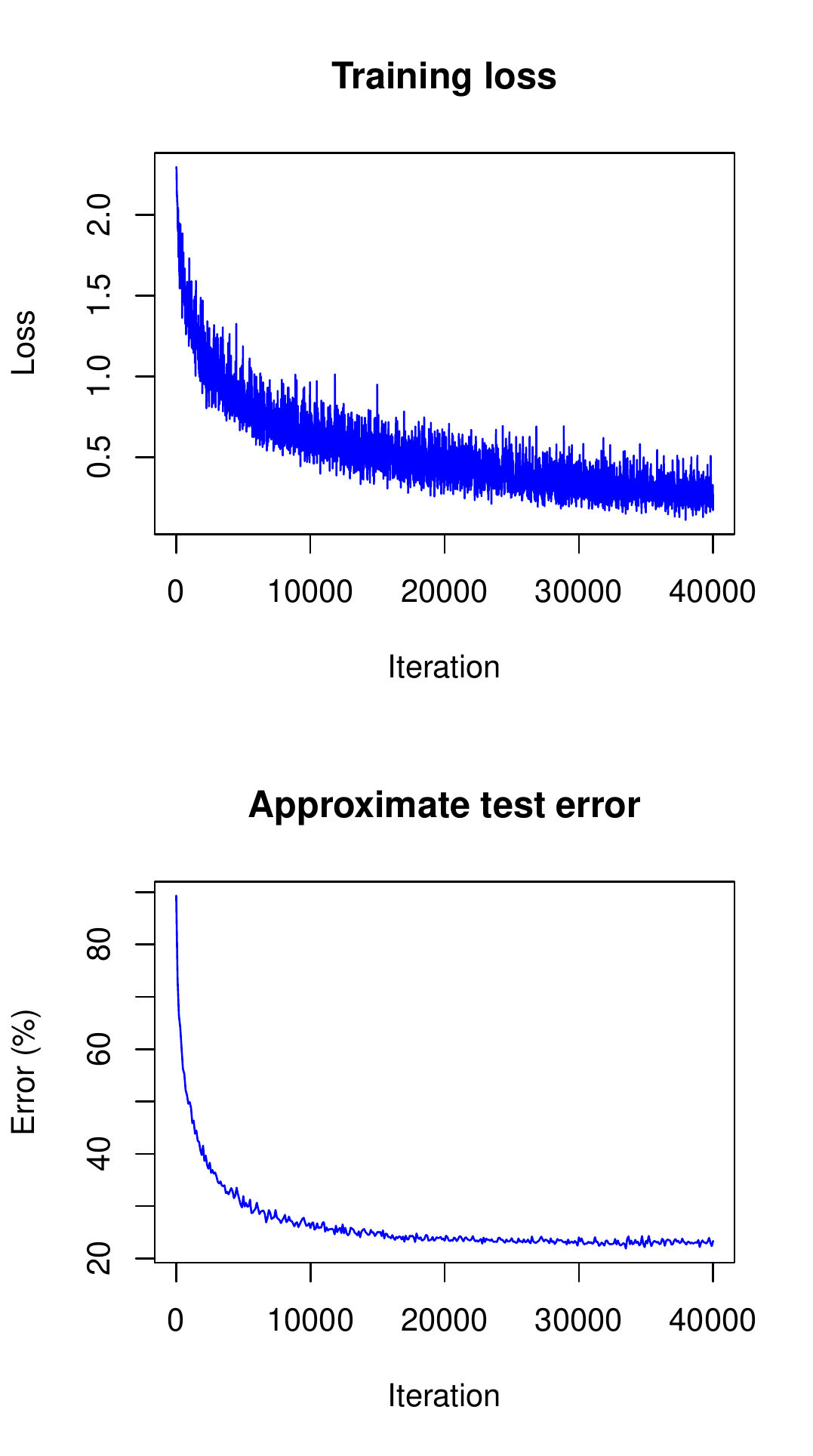}
\caption{Training visualization of the Bayesian model. For the computation of the approximate test error only one sample is drawn from the variational distribution per image.}  
\label{bayescifar10} 
    \end{minipage}  
\end{figure*}

The Figures \ref{freqcifar10} and \ref{bayescifar10} show the frequentist and the Bayesian training process, respectively. 
The achieved test errors are given in Table \ref{table:errorcifar10}. Note that the predictions of the Bayesian model are based on $200$ samples from the corresponding variational distribution per test image. The Bayesian model performs slightly better than the frequentist one due to the natural robustness of Bayesian models to overfitting.
\begin{table}
\caption{Test errors of the trained models. The predictions of the Bayesian model are based on $200$ samples from the corresponding variational distribution per test image.}
\label{table:errorcifar10}
\centering 
\begin{tabularx}{\linewidth}{ l@{\hskip 1.5in}  l }
  \hline			
  Model & Test error  \\
  \hline\\
  Frequentist  & 20.9\%  \\
  Bayesian  & 20.32\% \\
  \hline  
\end{tabularx}
\end{table}
In Table \ref{table:variationalcifar} the correlations $\rho_j$ and the variance determining parameters $\tau_j$ of the Bayesian model can be found ($j=1,...,4$). The parameters can be interpreted as in Section \ref{mnist}.
\begin{table}
\caption{Parameters $\rho_j$ and $\tau_j$ of the variational distribution ($j=1,...,4$).}
\label{table:variationalcifar}
\centering 
\small
\begin{tabularx}{\linewidth}{ l@{\hskip 0.9in}  l l l l }
  \hline		
  Layer  &  $\tau_j$ &  $\tau_{bj}$ &  $\rho_{j}$ &  $\rho_{bj}$ \\
  \hline\\
  Convolutional 1    & 0.04 & 0.05 & -0.47 & ~0.01\\
  Convolutional 2    & 0.46 & 0.05 & -0.49 & -0.01\\
  Convolutional 3    & 0.37 & 0.05 & -0.41 & ~0.01\\
  Fully connected    & 0.14 & 0.05 & -0.29 & ~0.01\\
  \hline  
\end{tabularx}
\end{table}
To quantify the quality of the prediction uncertainty of the Bayesian model we estimate the $\alpha=95\%$ credible intervals and further the $\alpha=99\%$ credible intervals of the component-wise network outputs for all test images. The estimates are based on $200$ random network outputs, respectively. The terms quite certain and uncertain are to be understood as in Section \ref{mnist}. Then Table \ref{table:preduncmnist} summarizes in how many of the correct classification results the model is certain about its prediction and further summarizes the same for the wrong classification results. The quality of the uncertainty information is good, since for the correct classifications the majority of the predictions is considered as quite certain in contrast to the incorrect classifications where the opposite is the case. The amount of uncertainty can be regulated by the parameter $\alpha$. Assigning a higher value to $\alpha$ results in higher uncertainty and thus also in a lower number of quite certainly deemed wrong predictions. Thus, an optimal choice of $\alpha$ is dependent on the application of interest. For instance, for self-driving cars a value of $\alpha$ near one would be appropriate. 
\begin{table}[!t]
\caption{Overview of quite certain and uncertain prediction results for $\alpha=95\%$ and $\alpha=99\%$.}
\label{table:predunccifar}
\centering 
\small
\begin{tabularx}{\linewidth}{l@{\hskip 0.25in} l@{\hskip 0.25in} l @{\hskip 0.25in}l  }
  \hline		
  Prediction result & $\alpha$ &  Quite certain &  Uncertain \\
  \hline\\
   correct & 95\% &5894 & 2074  \\
   wrong   & 95\% &403  & 1629   \\
   correct & 99\% &5281 & 2687  \\
   wrong   & 99\% &250  & 1782   \\
  \hline  
\end{tabularx}
\end{table}

\section{Conclusion}
\label{conclusion}

We presented the first Bayesian approach to deep learning that allows to learn correlations between network parameters while introducing only few additional parameters to be optimized. In particular, we approximated the intractable posterior of the network parameters with a product of Gaussian distributions with tridiagonal covariance matrices. These distributions are defined in such a way, that the variances are multiples of the expectation values and the correlations belonging to a given distribution are identical. The novel approach was evaluated on basis of the popular benchmark datasets MNIST and CIFAR-10. Superior prediction accuracies compared to well regularized frequentist models show that the proposed method is more robust to overfitting than classical ones. Further, we showed that accurate uncertainty information about network predictions can be computed. This possibility of measuring prediction uncertainty should have a significant impact for real world applications such as self-driving vehicles. Finally, network parameter uncertainties and dependencies can readily be interpreted per layer, due to the fact that only few additional parameters are required by our method.

\section*{Acknowledgments}
The publication is one of the results of the project iDev40 (www.idev40.eu). The iDev40 project  has  received  funding  from  the  ECSEL  Joint  Undertaking (JU) under  grant  agreement  No  783163. The JU receives support from the European Union’s Horizon  2020  research  and  innovation  programme. It  is  co-funded  by  the  consortium  members, grants from Austria, Germany, Belgium, Italy, Spain and Romania.
\vspace{-1cm}
\begin{figure}[H]
\centering
\begin{minipage}{.3\linewidth}
  \centering
  \includegraphics[width=\linewidth]{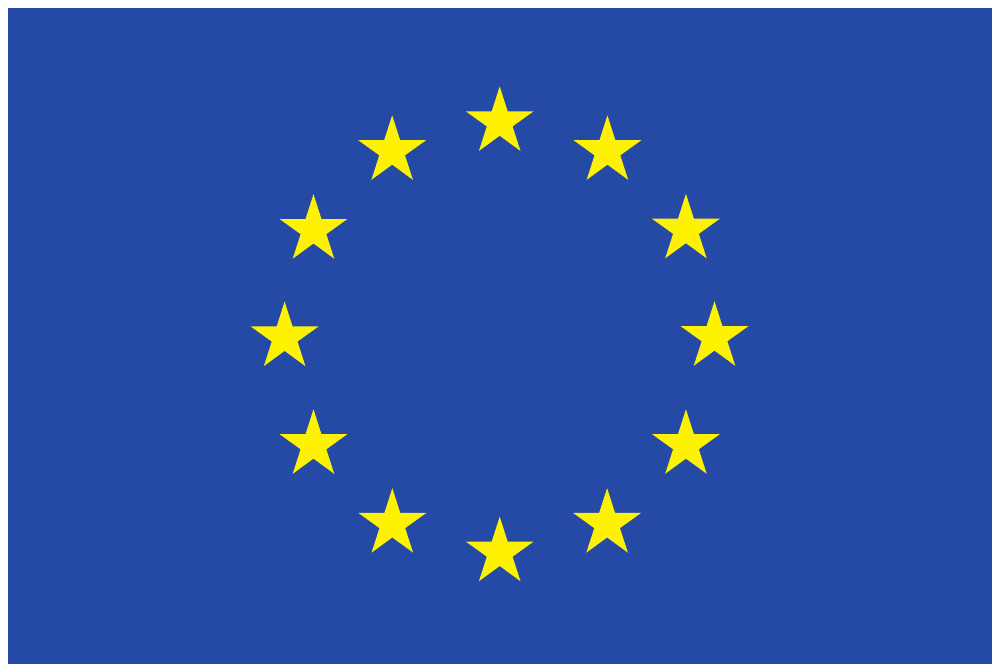}
\end{minipage}%
\begin{minipage}{.3\linewidth}
  \centering
  \includegraphics[width=\linewidth]{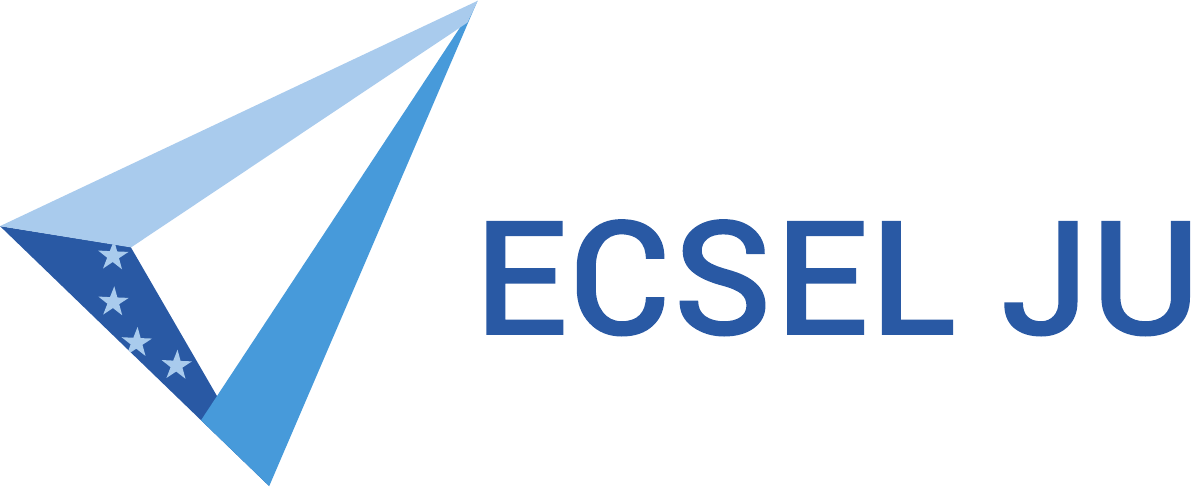}
\end{minipage}%
\hspace{.02\textwidth}
\begin{minipage}{.3\linewidth}
  \centering
  \includegraphics[width=\linewidth]{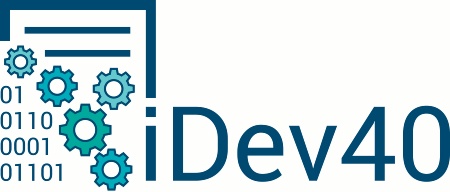}
\end{minipage}    
\end{figure}

\vspace{-1cm}

\ifCLASSOPTIONcaptionsoff
  \newpage
\fi

\IEEEtriggeratref{8}


\bibliographystyle{IEEEtran}
\bibliography{literature}

\begin{thebibliography}{10}
\providecommand{\url}[1]{#1}
\csname url@samestyle\endcsname
\providecommand{\newblock}{\relax}
\providecommand{\bibinfo}[2]{#2}
\providecommand{\BIBentrySTDinterwordspacing}{\spaceskip=0pt\relax}
\providecommand{\BIBentryALTinterwordstretchfactor}{4}
\providecommand{\BIBentryALTinterwordspacing}{\spaceskip=\fontdimen2\font plus
\BIBentryALTinterwordstretchfactor\fontdimen3\font minus
  \fontdimen4\font\relax}
\providecommand{\BIBforeignlanguage}[2]{{%
\expandafter\ifx\csname l@#1\endcsname\relax
\typeout{** WARNING: IEEEtran.bst: No hyphenation pattern has been}%
\typeout{** loaded for the language `#1'. Using the pattern for}%
\typeout{** the default language instead.}%
\else
\language=\csname l@#1\endcsname
\fi
#2}}
\providecommand{\BIBdecl}{\relax}
\BIBdecl

\bibitem{Bengio2006}
Y.~Bengio, H.~Schwenk, J.-S. Sen{\'e}cal, F.~Morin, and J.-L. Gauvain, ``Neural
  probabilistic language models,'' in \emph{Innovations in Machine Learning:
  Theory and Applications}, D.~E. Holmes and L.~C. Jain, Eds.\hskip 1em plus
  0.5em minus 0.4em\relax Springer, 2006, pp. 137--186.

\bibitem{Krizhevsky2012}
A.~Krizhevsky, I.~Sutskever, and G.~E.~Hinton, ``Imagenet classification with
  deep convolutional neural networks,'' \emph{Neural Information Processing
  Systems}, vol.~25, pp. 1097--1105, 2012.

\bibitem{Simonyan2014VeryDC}
K.~Simonyan and A.~Zisserman, ``Very deep convolutional networks for
  large-scale image recognition,'' vol. arXiv:1409.1556, 2014.

\bibitem{He2016DeepRL}
K.~He, X.~Zhang, S.~Ren, and J.~Sun, ``Deep residual learning for image
  recognition,'' \emph{2016 IEEE Conference on Computer Vision and Pattern
  Recognition (CVPR)}, pp. 770--778, 2016.

\bibitem{Hornik1990UniversalAO}
K.~Hornik, M.~B. Stinchcombe, and H.~White, ``Universal approximation of an
  unknown mapping and its derivatives using multilayer feedforward networks,''
  \emph{Neural Networks}, vol.~3, pp. 551--560, 1990.

\bibitem{Liang2016WhyDN}
S.~Liang and R.~Srikant, ``Why deep neural networks for function
  approximation?'' vol. arXiv:1610.04161, 2016.

\bibitem{Jozwik2017}
K.~Jozwik, N.~Kriegeskorte, K.~R~Storrs, and M.~Christina~Mur, ``Deep
  convolutional neural networks outperform feature-based but not categorical
  models in explaining object similarity judgments.'' in \emph{Front.
  Psychol.}, 2017.

\bibitem{Greenspan2016}
H.~Greenspan, B.~van Ginneken, and R.~M.~Summers, ``Guest editorial deep
  learning in medical imaging: Overview and future promise of an exciting new
  technique,'' \emph{IEEE Transactions on Medical Imaging}, vol.~35, pp.
  1153--1159, 2016.

\bibitem{Gulshan2016}
V.~Gulshan, L.~Peng, M.~Coram, M.~C.~Stumpe, D.~Wu, A.~Narayanaswamy,
  S.~Venugopalan, K.~Widner, T.~Madams, J.~Cuadros, R.~Kim, R.~Raman,
  P.~C.~Nelson, J.~L.~Mega, and D.~R.~Webster, ``Development and validation of
  a deep learning algorithm for detection of diabetic retinopathy in retinal
  fundus photographs,'' \emph{JAMA}, vol. 316, pp. 2402--2410, 2016.

\bibitem{Banerjee2017}
K.~Banerjee, T.~Van~Dinh, and L.~Levkova, ``Velocity estimation from monocular
  video for automotive applications using convolutional neural networks,'' in
  \emph{2017 IEEE Intelligent Vehicles Symposium (IV)}, 2017, pp. 373--378.

\bibitem{Heaton2016}
J.~B.~Heaton, N.~G.~Polson, and J.~H.~Witte, ``Deep learning for finance: deep
  portfolios,'' \emph{Applied Stochastic Models in Business and Industry},
  vol.~33, pp. 3--12, 2016.

\bibitem{Li2017}
X.~Li, Q.~Ding, and J.-Q. Sun, ``Remaining useful life estimation in
  prognostics using deep convolution neural networks,'' \emph{Reliability
  Engineering \& System Safety}, vol. 172, pp. 1--11, 2017.

\bibitem{dropout}
N.~Srivastava, G.~Hinton, A.~Krizhevsky, I.~Sutskever, and R.~Salakhutdinov,
  ``Dropout: A simple way to prevent neural networks from overfitting,''
  \emph{Journal of Machine Learning Research}, vol.~15, pp. 1929--1958, 2014.

\bibitem{dropconnect}
L.~Wan, M.~Zeiler, S.~Zhang, Y.~L. Cun, and R.~Fergus, ``Regularization of
  neural networks using dropconnect,'' in \emph{Proceedings of the 30th
  International Conference on Machine Learning}, vol.~28, 2013, pp. 1058--1066.

\bibitem{Gal2015orig}
Y.~Gal and Z.~Ghahramani, ``Bayesian convolutional neural networks with
  bernoulli approximate variational inference,'' vol. arXiv:1506.02158, 2015.

\bibitem{Kingma2015}
D.~P. Kingma, T.~Salimans, and M.~Welling, ``Variational dropout and the local
  reparameterization trick,'' in \emph{Advances in Neural Information
  Processing Systems 28}.\hskip 1em plus 0.5em minus 0.4em\relax Curran
  Associates, Inc., 2015, pp. 2575--2583.

\bibitem{Roman2017}
A.~Roman, ``Bayesian deep learning with edward (and a trick using dropout),''
  PyData London, 2017.

\bibitem{GalUnc}
Y.~Gal and Z.~Ghahramani, ``Dropout as a bayesian approximation: Representing
  model uncertainty in deep learning,'' \emph{Proceedings of The 33rd
  International Conference on Machine Learning}, pp. 1050--1059, 2015.

\bibitem{Buntine1991}
W.~Buntine and A.~S.~Weigend, ``Bayesian back-propagation,'' \emph{Complex
  Systems}, vol.~5, pp. 603--643, 1991.

\bibitem{Hinton1993KeepingTN}
G.~E. Hinton and D.~van Camp, ``Keeping the neural networks simple by
  minimizing the description length of the weights,'' in \emph{COLT}, 1993.

\bibitem{Denker91transformingneural-net}
J.~Denker and Y.~Lecun, ``Transforming neural-net output levels to probability
  distributions,'' in \emph{Advances in Neural Information Processing Systems
  3}.\hskip 1em plus 0.5em minus 0.4em\relax Morgan Kaufmann, 1991, pp.
  853--859.

\bibitem{MacKay1995}
D.~J.C.~MacKay, ``Probable networks and plausible predictions — a review of
  practical bayesian methods for supervised neural networks,'' \emph{Network:
  Computation in Neural Systems}, vol.~6, pp. 469--505, 1995.

\bibitem{neal2012bayesian}
R.~M. Neal, \emph{Bayesian learning for neural networks}.\hskip 1em plus 0.5em
  minus 0.4em\relax Springer Science \& Business Media, 2012, vol. 118.

\bibitem{Graves2011PracticalVI}
A.~Graves, ``Practical variational inference for neural networks,'' in
  \emph{NIPS}, 2011.

\bibitem{Blundell:2015:WUN:3045118.3045290}
C.~Blundell, J.~Cornebise, K.~Kavukcuoglu, and D.~Wierstra, ``Weight
  uncertainty in neural networks,'' in \emph{Proceedings of the 32Nd
  International Conference on International Conference on Machine Learning -
  Volume 37}, 2015, pp. 1613--1622.

\bibitem{Louizos2016}
C.~Louizos and M.~Welling, ``Structured and efficient variational deep learning
  with matrix gaussian posteriors,'' in \emph{Proceedings of the 33rd
  International Conference on International Conference on Machine Learning -
  Volume 48}, 2016, pp. 1708--1716.

\bibitem{Amari1985}
S.-i. Amari, \emph{Differential-geometrical methods in statistics}.\hskip 1em
  plus 0.5em minus 0.4em\relax Springer, 1985.

\bibitem{kullback1951}
S.~Kullback and R.~A. Leibler, ``On information and sufficiency,'' \emph{The
  Annals of Mathematical Statistics}, vol.~22, pp. 79--86, 1951.

\bibitem{Hernndez2016}
J.~M. Hern{\'a}ndez-Lobato, Y.~Li, M.~Rowland, T.~D. Bui,
  D.~Hern{\'a}ndez-Lobato, and R.~E. Turner, ``Black-box alpha divergence
  minimization,'' in \emph{ICML}, 2016.

\bibitem{GalAlpha2017}
Y.~Li and Y.~Gal, ``Dropout inference in bayesian neural networks with
  alpha-divergences,'' in \emph{ICML}, 2017.

\bibitem{Posch2019}
K.~{Posch}, J.~{Steinbrener}, and J.~{Pilz}, ``{Variational Inference to
  Measure Model Uncertainty in Deep Neural Networks},'' vol. arXiv:1902.10189,
  Feb 2019.

\bibitem{Deng2012}
L.~Deng, ``The mnist database of handwritten digit images for machine learning
  research,'' \emph{IEEE Signal Processing Magazine}, vol.~29, pp. 141--142,
  2012.

\bibitem{Krizhevsky2009LearningML}
A.~Krizhevsky, ``Learning multiple layers of features from tiny images,'' 2009.

\bibitem{Bishop2006}
C.~M. Bishop, \emph{Pattern Recognition and Machine Learning (Information
  Science and Statistics)}.\hskip 1em plus 0.5em minus 0.4em\relax Springer,
  2006.

\bibitem{Andelic2011}
M.~Andeli{\'{c}} and C.~Fonseca, ``Sufficient conditions for positive
  definiteness of tridiagonal matrices revisited,'' \emph{Positivity}, vol.~15,
  pp. 155--159, 2011.

\bibitem{Hershey2007}
J.~R. Hershey and P.~A. Olsen, ``Approximating the kullback leibler divergence
  between gaussian mixture models,'' \emph{2007 IEEE International Conference
  on Acoustics, Speech and Signal Processing}, vol.~4, pp. IV--317--IV--320,
  2007.

\bibitem{jia2014caffe}
Y.~Jia, E.~Shelhamer, J.~Donahue, S.~Karayev, J.~Long, R.~Girshick,
  S.~Guadarrama, and T.~Darrell, ``Caffe: Convolutional architecture for fast
  feature embedding,'' \emph{arXiv:1408.5093}, 2014.

\bibitem{Lecun1998}
Y.~Lecun, L.~Bottou, Y.~Bengio, and P.~Haffner, ``Gradient-based learning
  applied to document recognition,'' \emph{Proceedings of the IEEE}, vol.~86,
  pp. 2278 -- 2324, 1998.

\end{thebibliography}
%



%

\vspace{-0.3cm}
\begin{IEEEbiography}[{\includegraphics[width=1in,height=1.25in,clip,keepaspectratio]{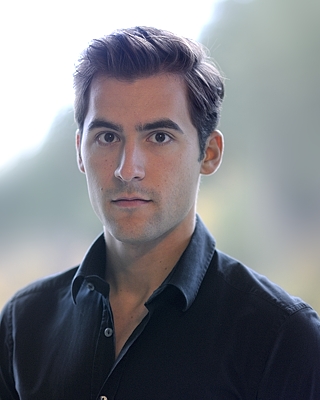}}]{Konstantin Posch}
Konstantin Posch obtained his B.Sc. and his M.Sc. in Technical Mathematics from Alpen- Adria- University Klagenfurt in 2016 and 2018. Currently, he is working on his PhD in Technical Mathematics at University Klagenfurt. His research interests include Bayesian Statistics, Statistical Machine Learning and Big Data Analytics.
\end{IEEEbiography}
\begin{IEEEbiography}[{\includegraphics[width=1in,height=1.25in,clip,keepaspectratio]{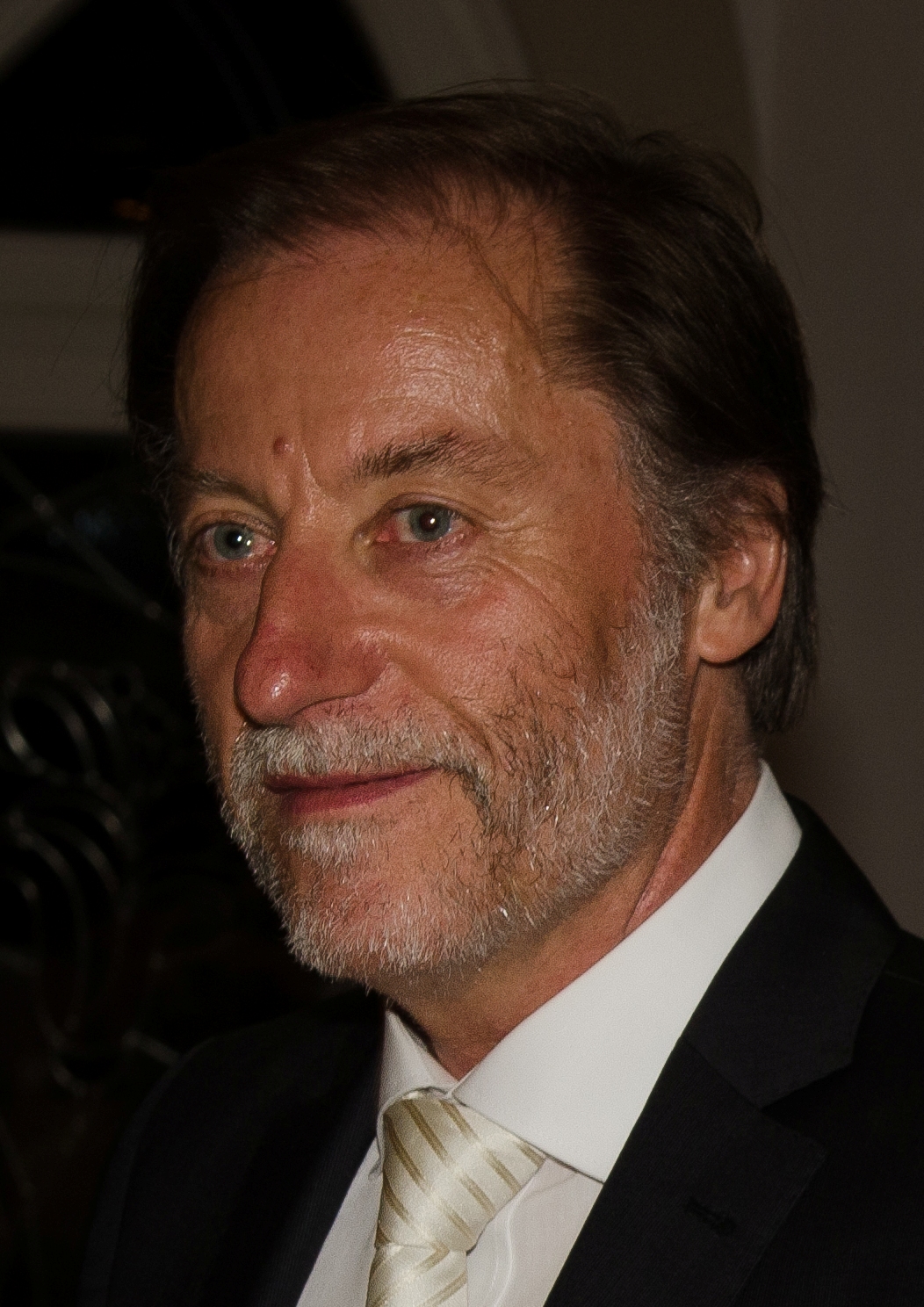}}]{Juergen Pilz}
Juergen Pilz is Professor of Applied Statistics at the Alpen- Adria- University
(AAU) of Klagenfurt, Austria. He has authored more than 150 publications
in international journals and conference proceedings in the areas of Bayesian
statistics, spatial statistics, environmental and industrial statistics, statistical
quality control and design of experiments. He is the author of seven books
which appeared in internationally renowned publishing houses such as
Wiley, Springer and Chapman and Hall, and authored several book chapters.
He obtained his M.Sc., PhD and D.Sc. degrees from Freiberg Technical University
in Germany in 1974, 1978 and 1988, respectively, in the areas of mathematics
and statistics. Since 2007 he serves as head of the Department of Statistics at
AAU Klagenfurt, Austria. He is elected member of the International Statistical
Institute (ISI) and the Institute of Mathematical Statistics (IMS). He is on editorial
boards of several international statistics journals.
\end{IEEEbiography}




\end{document}